\documentclass[]{analogyai}
\usepackage{amsmath}
\usepackage{amsfonts}
\usepackage{textcomp}
\usepackage{tabularx}
\usepackage{booktabs}
\usepackage{enumitem}
\usepackage{lipsum}

\title{FM-Bench: A Benchmark for Long-Horizon Management
with Competing Agents}

\author[1,*]{Tianyou Wang}
\author[1,*]{Chongyang Gao}
\author[1]{Kezhen Chen}
\author[1]{Dong Chen}
\author[1]{Yinghao He}
\author[1]{Donghan Li}
\author[1]{Wangcheng Xu}
\author[1]{Yifan Xiao}
\author[1]{Hongjiu Zhang}
\author[1]{Chi Li}

\affiliation[1]{AnalogyAI}

\contribution[*]{Equal contribution}

\abstract{Language model agents now execute bounded tasks reliably. Whether they can sustain effective decision-making over long horizons, where actions have cumulative consequences and the environment responds to their choices, remains largely unmeasured. FM-Bench (Football Management Benchmark) measures this. An LLM agent runs a football club for 20 in-game years through 26 tools and roughly 340 to 400 decision stops, each admitting as many tool calls as the agent chooses to spend. It drafts a squad on the same fixed budget as every rival, trades players, negotiates contracts, invests in facilities and youth, sets lineups, and answers to a board that can fire it. A deterministic engine grades the run, accumulating across years into one final score with no LLM judge or human rater. Evaluation runs in two modes. The solo track plays each of 15 frontier models against a frozen scripted world, and the Arena places the same models plus a scripted anchor in one shared 20-year world; to our knowledge, the first head-to-head evaluation at this scale and horizon. We measure six behavioral capabilities behind its score. Across three seeds, all 15 models complete every horizon while the blind scripted baselines die out in most of theirs, and \texttt{claude-fable-5} tops the solo board on mean score and the Arena, where the league title nonetheless rotates among ten models. Neither scale, price, nor vendor predicts the order; the order settles only late in the horizon, and the best first-play human lands only at the bottom of the model board. What separates the models is managerial behavior rather than computation. Higher-scoring models reduce slow-payoff investment as the horizon approaches its end, keep cash invested rather than idle, and open contract renewals well before the deadline, while token spend predicts nothing. Furthermore, no model learns the market's hidden prices from hundreds of rejected bids, and self-managed memory fails in two opposite modes: an archive that only grows or a plan rewritten every season. }

\date{2026-08-18}
\correspondence{Chi Li at \email{chili@analogyai.org}}
\metadata[Code]{\url{https://github.com/Analogy-AI/fm-bench}}

\begin{document}

\maketitle

\section{Introduction}\label{sec:intro}

Language model agents now handle bounded tasks reliably: they resolve real GitHub issues~\citep{jimenez2024swe}, solve competition-grade reasoning problems~\citep{cfe, lightman2024let, balunovic2026matharena}, complete computer-use workflows~\citep{zhou2024webarena, xie2024osworld}, and orchestrate multi-step tool calls~\citep{liu2024agentbench, qin2024toolllm, yao2024tau}. These tasks have a short horizon to the correct answer and no competing agents. Newer benchmarks push one dimension of realism, the long horizon, stretching execution to hundreds of steps~\citep{li2026long}, to reactive game worlds~\citep{kuttler2020nethack, hafner2021benchmarking, cote2018textworld, paglieri2025balrog}, and to business simulations that run a vending machine or shop for simulated months~\citep{backlund2025vendingbench, shi2026merchantbench}, a startup for a year~\citep{he2026yc, ceo}, or a company's finances through an eleven-year macro cycle~\citep{cfo}. Others push the second dimension, competing agents, but stage cooperation and competition over short episodes~\citep{zhu2025multiagentbench, wang2024battleagentbench} or rank models by bluffing and bargaining inside a single session~\citep{muller2026cattle}.

However, in real-world scenarios, these benchmarks still fall short of combining both dimensions at once and of providing an organization to keep alive through them. This combination is what we call \emph{management}, running an organization against rivals pursuing the same opportunities, over a horizon so long that early decisions reshape the world, with performance judged cumulatively under the conditions those decisions created. Management in this sense makes four demands on an agent. \emph{Hidden information}: the state that matters is never fully observable, so valuation must stay calibrated against permanent noise. \emph{Cumulative consequences}: the best actions pay off years later, losing positions compound yet remain recoverable, and honors accrue into the final score each season. \emph{A counter-adaptive environment}: other actors respond to revealed behavior, so no fixed strategy stays a best response and a winning move decays with use. \emph{Multi-objective pressure}: results and financial discipline are judged jointly, and satisfying one constraint can violate another.

\begin{figure}[t]
\centering
\includegraphics[width=\textwidth]{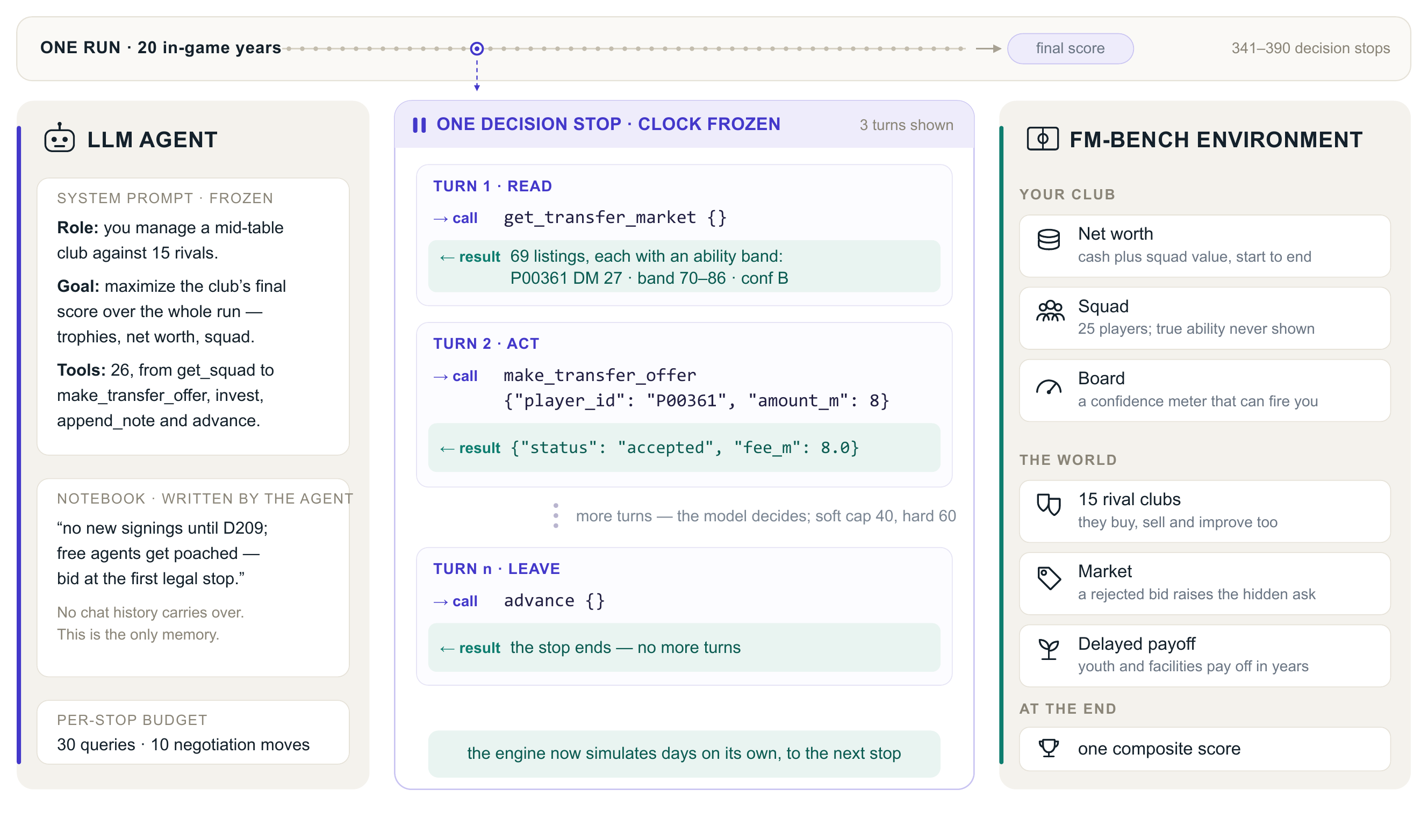}
\caption{How a run works. One run spans 20 in-game years and roughly 340 to 400
decision stops, ending in a single composite score. At each stop, the clock
freezes and the agent takes as many tool-call turns as it wants (read
$\to$ act $\to$ \texttt{advance}). The agent's only cross-stop memory is its self-written
notebook (\S\ref{sec:memory}); the environment side holds the club (net
worth, squad with permanently hidden true ability, a board that can fire
the manager) and the world (15 rival clubs, a counter-adaptive market,
delayed-payoff investments).}
\label{fig:overview}
\end{figure}

To measure management so defined, we present FM-Bench (Football Management Benchmark), where an LLM agent runs a football club for 20 in-game years under all four demands instantiated by concrete game mechanisms (\S\ref{sec:loop}, Figure~\ref{fig:overview}). Through 26 schema-generated tools over roughly 340 to 400 decision stops, it drafts a squad on the same fixed budget as every rival, trades players, negotiates contracts, invests in facilities and youth, fields lineups, and answers to a board that can fire it, while a deterministic engine accumulates every quantity into one final score with no LLM judge or human rater. Each stop is a fresh conversation whose only carried state is a self-authored notebook, so memory curation is itself measured. The \emph{solo track} plays each of 15 flagship models against a frozen scripted world with tiered opponents. The \emph{Arena} places all 16 seats in one shared 20-year world where every signing denies a rival the same player, made a valid measurement by an equal-endowment draft, sealed-bid conflict resolution, and capped revival. Across three seeds, every model completes the horizon, while the blind scripted anchors, including a disciplined heuristic, die out in 7 of their 9 runs. \texttt{claude-fable-5} tops both tracks, reaching about 95\% of a scripted upper anchor allowed to read the hidden state (90.94 against 95.54), yet competition still reshuffles the board, with the league title rotating among ten models and mid-board solo standings not surviving adaptive rivals. We decompose the scalar score into six behavioral capabilities, of which reducing slow-payoff investment near the horizon (Spearman $-0.58$), keeping cash deployed rather than idle ($-0.50$), and opening contract renewals early ($+0.45$) track the score on every seed, while token spend is uncorrelated under every accounting. Hundreds of rejected bids never teach a model where the market's true prices lie, and notebooks fail in two opposite regimes, append-only archives and wholesale rewrites.

Our contributions:
\begin{enumerate}[leftmargin=2em]
\item \textbf{FM-Bench, a benchmark for long-horizon management with
  competing agents.} A deterministic 20-year football-management
  environment that instantiates all four demands: hidden information,
  cumulative consequences, a counter-adaptive market, and
  multi-objective pressure. Player ability stays permanently hidden
  behind biased scouting, consequences accumulate across roughly 340 to 400
  decision stops with every season's honors accruing into the final
  score, the transfer market adapts against revealed strategies, and a
  board judges results and financial discipline jointly. Grading is
  continuous and cumulative, computed by the mechanics at every stop,
  with no LLM judge or human rater anywhere.
\item \textbf{Two tracks on one engine, isolating competition.} The
  solo track plays each model against a frozen scripted world, and the
  Arena, to our knowledge the first evaluation to run 15 frontier
  models as competing agents in one persistent economy over a 20-year
  horizon, places the same models in one shared world. Six first-play
  humans ran the same track; four died out and the best finished at the
  bottom of the model board. Their strengths mirror the models'
  weaknesses, making human--agent collaboration on this track a
  measurable, promising future direction.
\item \textbf{Long-horizon behavioral analysis.} We turn the scalar
  into a per-model profile over six behavioral capabilities. Strong
  models reduce slow-payoff investment before the horizon ends, keep
  cash invested rather than idle, and open contract renewals ahead of
  the deadline; hundreds of rejected bids never teach a model the
  market's true prices, notebooks fail as append-only archives or
  wholesale rewrites, and token spend predicts nothing. The profile
  explains what a leaderboard cannot. The winner is a generalist, the
  mid-board pairs one strength with one decisive gap, and the bottom
  fails on several axes at once.
\end{enumerate}

\section{Related Work}\label{sec:related}

\paragraph{Bounded tasks.} General agent benchmarks such as AgentBench~\citep{liu2024agentbench} and tool-orchestration suites such as ToolBench~\citep{qin2024toolllm} and $\tau$-bench~\citep{yao2024tau} measure whether a model can select and sequence tool calls correctly. WebArena~\citep{zhou2024webarena} and OSWorld~\citep{xie2024osworld} ground the same question in realistic software, where the difficulty is largely perceptual. SWE-bench~\citep{jimenez2024swe} and GAIA~\citep{mialon2024gaia} grade one-shot deliverables. In all of these, episodes are short, the environment is stationary, and mistakes rarely compound. FM-Bench keeps the same tool-calling interaction shape but removes the grounding burden and stretches the horizon to hundreds of decision stops in which early errors compound for years.

\paragraph{Long-horizon evaluation.} Terminal-agent marathons~\citep{li2026long, cheng2026terminal}, UltraHorizon~\citep{ultrahorizon2025}, and game environments~\citep{kuttler2020nethack, hafner2021benchmarking, cote2018textworld, paglieri2025balrog} stretch episodes to hundreds of steps, but the former pursue a fixed deliverable in a stationary world and the latter reward reactive skill rather than economic planning. Closest in spirit are business simulations that run a vending machine~\citep{backlund2025vendingbench}, an online shop~\citep{shi2026merchantbench}, a startup~\citep{ceo, he2026yc}, or a company's finances~\citep{cfo} for months to years of simulated time; YC-Bench's strongest predictor of success, scratchpad use, matches the memory-curation capability of \S\ref{sec:capabilities}. Even there, the environment does not push back, as StockBench replays real markets the agent's actions never move~\citep{chen2025stockbench} and CoffeeBench tests one model at a time against fixed reference agents~\citep{coffeebench2026}. FM-Bench differs in combining a counter-adaptive environment, permanently hidden state, an explicit memory discipline test, and cumulative mechanism-computed scoring.

\paragraph{Multi-agent evaluation and arenas.} Arena-style rankings began with pairwise human-preference voting~\citep{pmlr-v235-chiang24b}, where the models never interact, and coordination benchmarks~\citep{zhu2025multiagentbench, wang2024battleagentbench} measure how well agent systems cooperate over short episodes rather than who manages better over time. Competition between models appears in Cattle Trade, single-session negotiation ranked by TrueSkill with no organization to run~\citep{muller2026cattle}, and in Vending-Bench Arena, three models in a shared vending market documented only through a leaderboard page over a closed environment~\citep{andonlabs2026arena}. FM-Bench's Arena runs this line at a different scale, 15 frontier models plus a scripted anchor managing rival clubs for 20 seasons, with a matched solo track measuring the same model without live competition. Orthogonally, AgentAtlas~\citep{agentatlas2026} argues that outcome leaderboards conflate distinct competences and calls for the behavior-level decomposition that \S\ref{sec:capabilities} provides.

\begin{table}[t]
\centering
\small
\caption{The four demands: the mechanism that instantiates
each, the ability it targets, and the calibration lever
that tunes it.}
\label{tab:pillars}
\begin{tabularx}{\textwidth}{@{}>{\raggedright\arraybackslash}p{2.7cm}X>{\raggedright\arraybackslash}p{2.9cm}>{\raggedright\arraybackslash}p{2.5cm}@{}}
\toprule
demand & mechanism & targeted ability & calibration lever \\
\midrule
\textbf{Hidden information} & scout bands with a permanent per-scout
bias; hidden player traits (injury proneness, development rate, aging
onset); hidden asks in negotiation & valuation calibration under
noise & bandwidth, bias sigma \\
\addlinespace
\textbf{Cumulative consequences} & Upside: youth development and
facility investment pay off over years. Downside: insolvency ends in
administration (points penalty, forced sales), confidence collapse ends
in firing; both compound yet remain recoverable. Honors accrue into the
final score season by season & long-range credit assignment; trend
recognition, loss-cutting & growth and aging curves; spiral thresholds,
warning cadence; honors weights \\
\addlinespace
\textbf{Counter-adaptive market} & rejected bids raise the hidden ask;
repeat-pair markups (bargains included); anti-inversion counters;
negotiation cooldowns; per-seed mispricing & strategy adaptation & markup
and decay rates \\
\addlinespace
\textbf{Multi-objective pressure} & the board judges results and
financial discipline jointly; season targets scale with squad
strength & multi-constraint balancing & target cushion, board
patience \\
\bottomrule
\end{tabularx}
\end{table}

\section{FM-Bench}\label{sec:game}

This section describes the task and the interaction loop, the information design and the four demands it instantiates (\S\ref{sec:loop}), the memory design that turns curation into a measured capability (\S\ref{sec:memory}), and the score (\S\ref{sec:scoring}).

\subsection{Environment Overview}\label{sec:loop}

\paragraph{The task.} The agent manages one club in a 16-club professional division. Every club, the agent and all rivals alike, drafts its 25-player squad from the same template pool under the same fixed budget, and picks are independent copies, so no club's draft interferes with another's. From there, the agent runs the club for 20 in-game years: trading players, negotiating contracts, investing in facilities and youth, setting the lineups and tactics that decide its matches, and answering to a board that can fire it. A run ends in a single composite score that accumulates season by season (\S\ref{sec:scoring}). Firing or insolvency does not end measurement but triggers capped revival at a score discount.

\paragraph{A deterministic world.}Everything is generated deterministically from the seed, including the draft pool itself. The only non-seed inputs are the agents' own decisions, of which the draft picks are the first. The seed fixes one division, double round-robin with 30 rounds per season on a 360-day calendar, ${\sim}470$ fully simulated players alive at any time, the $16 \times 25$ senior rosters plus youth academies, and ${\sim}2{,}000$ distinct players over a 20-year run as youth intakes arrive and veterans retire. No real-world names or data appear anywhere, so memorized football knowledge cannot leak in.

\paragraph{The interaction loop and stop.} The engine simulates day-by-day and pauses only at \emph{decision stops}: 13 scheduled stops on a fixed season calendar, plus event-driven stops the world raises unsolicited, ${\sim}6$ per season for ${\approx}19$ stops per season in total. Completed 20-year runs log roughly 340-400 stops, with the exact count varying with play. Several types can fire at once, and a stop carries the full list of reasons the world stopped (Table~\ref{tab:stops}, Appendix~\ref{app:tools}). Every stop delivers the same packet: the list of stop types, a one-glance digest (league position, points, season target, cash, wage ratio, board confidence, injuries, next fixture, open window, and any active warnings), the inbox of everything that happened since the last stop (19 unsolicited event types, from incoming transfer offers and counter-bids to retirements and completed facilities) with action-required items flagged, live pending offers, the agent's last 10 state-changing actions, and its notebook (\S\ref{sec:memory}). The agent then takes as many tool-call turns as it wants, using query tools to inspect anything and action tools to decide, in any order, and finally calls \texttt{advance} to move time forward. The full 26-tool interface is listed in Table~\ref{tab:tools}, Appendix~\ref{app:tools}. Inaction is conservative, not fatal: ignored offers expire, lineups and tactics persist, open negotiations run until offers lapse after 10 game days, and up to 10 single-condition standing orders such as ``accept offers above $2\times$ value'' execute autonomously between stops, deliberately too weak to replace per-stop judgment. Figure~\ref{fig:overview} shows the run / decision-stop / turn structure at a glance. Appendix~\ref{app:turn} reproduces one stop verbatim as the agent sees it, and Table~\ref{tab:scale} in Appendix~\ref{app:tools} gives the scale of one 20-year run by the numbers.

\paragraph{The four demands.} The four demands that define management in \S\ref{sec:intro} are not prompt dressing; each is a concrete mechanism with a targeted ability and a calibration lever (Table~\ref{tab:pillars}). \emph{Hidden information}: a player's true ability, potential, and traits (injury proneness, development rate, aging onset, consistency, style fit) are revealed to no one, the owning club included. Scouting returns bands with a permanent per-scout bias, so estimates converge to ``truth + your scout's bias'' and trait estimation is statistical inference over seasons, not a lookup, while season stats, contracts, and finances stay exact. \emph{Cumulative consequences}: youth development and facility investment pay off years later, insolvency and confidence collapse compound toward administration and firing yet remain recoverable, and honors enter the final score season by season. \emph{Counter-adaptive market}: the market reacts to revealed behavior, rejected bids raise the hidden ask, repeated dealings with the same club raise its prices, and negotiation spam triggers cooldowns, so no fixed strategy stays a best response. \emph{Multi-objective pressure}: the board judges results and financial discipline jointly, against season targets that scale with squad strength. Recorded runs show models failing against each demand. A mid-tier model let a player talk its wage offer up step by step, from 12 to 14 to 17 to 18M per year, accepting each demanded raise at face value because it could not judge what the player was truly worth (demand 1), and the same model went insolvent with wages above revenue (demand 2). Every model under-bid a market that raises its hidden ask after each rejection, needing a median of 30 offers per completed signing (demand 3, \S\ref{sec:capabilities}). A small model was fired by its board in every run (demand 4).

\subsection{Memory Design}\label{sec:memory}

Memory in FM-Bench is the agent's responsibility, not a harness feature. Every decision stop opens as a fresh conversation with no chat history. Whatever the agent wants its future self to know, it must write down, so memory curation becomes part of what the benchmark measures. Four layers supply what the agent ``remembers'' (Table~\ref{tab:memory}, Appendix~\ref{app:tools}): the \emph{packet auto-context}, pushed by the engine into every stop for free, covering current league position, cash, board confidence, injuries, offers on the table, and everything that happened since the last stop; the \emph{archive}, past facts queryable on demand, from any season's final table to every honor and transfer of the run; the \emph{audit log}, which records every action for replay and anti-cheat but is never shown to the agent; and the \emph{notebook}, the only carrier of intent, holding the agent's own plans and reasons such as ``why I bought him'' or ``sell Y in winter'', written by the agent and injected back into every packet.

We chose this design for three reasons. \emph{Scale}: a 20-year run generates far more history than any context window holds, so some forgetting policy is unavoidable. A rolling-context harness would impose one implicitly and untraceably, while the notebook makes it explicit, bounded, and the agent's own. \emph{Comparability}: if the harness curated context by truncation, summarization, or retrieval over an external store, scores would partly measure the harness. Here every model faces the same contract, and what differs between models is only what they chose to write. \emph{Measurement}: deciding what the future self will need is itself part of the capability under test. The notebook then becomes analysis data, and the memory-curation metrics and the notebook audit of \S\ref{sec:capabilities} are read directly off it.

\subsection{Scoring}\label{sec:scoring}

The score is built so that every season of a 20-year run counts. It
aggregates quantities the game mechanics compute at every stop, honors
earned in each season enter the final score directly, and no phase can
be written off or made up by an endgame sprint. Notation: $H$ =
cumulative honors points; $\mathit{VA}$ = net-worth \emph{value added};
$M$ = squad value; $\rho$ = early-settlement discount. The score has a
single source-of-truth implementation in the engine:
\begin{align*}
S_{\text{raw}} \;=\;
\underbrace{18\,\ln\!\Bigl(1 + \tfrac{\max(H,\,0)}{60}\Bigr)}_{\text{honors}}
\;+\;
\underbrace{10\,\operatorname{sign}(\mathit{VA})\,
\ln\!\Bigl(1 + \tfrac{|\mathit{VA}|}{40}\Bigr)}_{\text{value added (primary)}}
\;+\;
\underbrace{6\,\ln\!\Bigl(1 + \tfrac{\max(M,\,0)}{80}\Bigr)}_{\text{squad value (stabilizer)}}
\\[10pt]
S_{\text{final}} \;=\; \rho \cdot \max(S_{\text{raw}},\,0)
  + \min(S_{\text{raw}},\,0)
\end{align*}

\begin{itemize}[leftmargin=2em]
\item $H$ sums honors points over the whole run: champion $+100$, top-4
  $+20$. Penalties offset honors inside the sum, but the
  channel argument is floored at zero, so honors debt cannot drive the
  term negative.
\item $\mathit{VA} = W_{\text{real}} - W_{\text{baseline}}$, in M credits:
  $W_{\text{real}}$ is the mean of the last three seasons' net-worth
  snapshots, deflated at 4\%/yr, and $W_{\text{baseline}}$ is the
  \emph{same seed's} day-0 net worth. Zero reads ``you handed the club back as
  valuable as you found it, in real terms.'' Net worth books cash, plus
  squad value at a 0.8 liquidity discount, plus facilities at 0.5 while
  under construction, minus outstanding transfer-fee liabilities. Cash
  above $3\times$ the annual wage bill counts at only 0.3, an
  anti-hoarding discount.
\item $M$ is the last-3-season deflated mean squad value, a low-weight
  stabilizer so that a cash-hoarding squad-stripper cannot win on net worth
  alone.
\item $\rho = 1$ for a completed run. For a run that settles early, by
  firing, insolvency, or abandonment, $\rho = 0.8 \cdot (t/T)^{1.5}$ and
  multiplies only the positive part, so dying early can never shrink a
  negative result, and there is no fire-to-cap-losses arbitrage. Humans
  abandoning mid-run settle by the same rule.
\end{itemize}

All three channels are log-compressed and uncapped, so no channel can run away, and the benchmark cannot saturate. The weights and divisors are calibration constants, not derived quantities. They were tuned on the free scripted anchors and frozen before any scored run, and every adjustment was accepted only if four criteria held. \emph{Order preservation}: the anchor ladder stays monotonic, random $<$ idle $<$ heuristic $<$ oracle, where the heuristic is a disciplined hand-written management script, and the oracle is the privileged truth-reading script of \S\ref{sec:protocol}. \emph{Discrimination}: the gap between disciplined and passive anchors stays wide. \emph{Floor}: the oracle ceiling stays positive on every seed. \emph{Non-saturation}: top scores keep separating instead of bunching at a ceiling. Appendix~\ref{app:score} states and proves the basic shape properties of the composite: no fire-to-cap arbitrage, channel monotonicity, fair-price transfer neutrality on the VA channel, and diminishing returns.

\section{Experiments and Results}\label{sec:experiments}

\subsection{Evaluation Settings}\label{sec:protocol}

\paragraph{Configuration.} Every club starts as an identical shell with reputation 70, a 38{,}000-seat stadium, and a 100M-credit draft budget over the shared 50-template pool, and unspent budget plus a 50M stipend becomes opening cash. An episode spans 20 in-game years of 360 days each, one division of 16 clubs playing 30 rounds per season, and roughly 340 to 400 decision stops. All results in this paper come from this 20-year track on the tiered world, the solo runs on seeds 1, 1024, and 2026, and the Arena on seed 7, and the six human runs (\S\ref{sec:campaign}) play the same track. Scores are comparable only within a track and calibration version. Every result file is stamped with the engine commit, params hash, score version, and the full knob set, and the ladder report warns loudly on mixed pools.

\paragraph{Models and harness.} The roster covers 15 flagship models plus one scripted anchor. Ten are closed frontier models, four from Anthropic (\texttt{claude-fable-5}, \texttt{claude-opus-4.8}, \texttt{claude-sonnet-5}, \texttt{claude-haiku-4.5}), two from OpenAI (\texttt{gpt-5.6-sol}, \texttt{gpt-5.6-terra}), two from Google (\texttt{gemini-3.5-flash}, \texttt{gemini-3-flash}), \texttt{grok-4.5} from xAI, and \texttt{muse-spark-1.1} from Meta. Five are open-weight models (\texttt{deepseek-v4-pro}, \texttt{kimi-k2.6}, \texttt{qwen3.7-max}, \texttt{glm-5.2}, \texttt{minimax-m3}). The selection fixes within-family tier curves, closed against open frontier, and an agentic specialist (Table~\ref{tab:roster}, Appendix~\ref{app:roster}). Every model runs under our fixed harness with locked model ID, temperature, and prompt, at the provider's recommended full reasoning configuration, and the harness never handicaps a model's reasoning to save cost. The effective setting is stamped into every result, including any degradation to no-thinking, so reasoning-on and reasoning-off pools can never merge silently. Our runs are reasoning-on, and no custom harness is allowed in either mode.

\paragraph{Modes.} The solo track plays one tested LLM against 15 scripted clubs and reports absolute $S_{\text{final}}$. The Arena places the 15 LLMs plus the scripted anchor in one shared world and ranks them by the same $S_{\text{final}}$. The equal-endowment draft, sealed-bid resolution, and capped revival remove seat confounds by mechanism, so a single shared run is valid on its own, and replication comes from seeds (\S\ref{sec:arena}). Every number in this paper comes from one stamped run per seat, with no run selection.

\paragraph{The oracle ceiling.} The oracle is a scripted policy whose privilege is information, not power. It reads true hidden state directly from the engine, but every action still goes through the same 26-tool interface, per-stop budgets, and legality checks as any agent. It converts that access into exact transfer prices, exact reservation wages, true-potential academy picks, peak-timed sales, and a starting eleven re-solved from true ability, plus a survival controller that keeps it from ever being fired (Appendix~\ref{app:oracle}). Being a script rather than an optimum, the ceiling is soft in principle.

\begin{table}[t]
\centering
\small
\caption{The Solo track results. The four non-LLM seats are reference policies. The oracle reads
true hidden state through the same interface (\S\ref{sec:protocol},
Appendix~\ref{app:oracle}); heuristic is a disciplined blind script,
idle never acts, and random acts arbitrarily
(Appendix~\ref{app:anchors}). Both columns are mean $\pm$ sd over
seeds 1 / 1024 / 2026; tokens are the run total by manifest
accounting, in millions.}
\label{tab:campaign}
\begin{tabular}{@{}lrr@{}}
\toprule
seat & $S_{\text{final}}$ & tokens (M) \\
\midrule
oracle (privileged) & $95.54 \pm 4.68$ & --- \\
\texttt{claude-fable-5} & $90.94 \pm 5.20$ & $31 \pm 5$ \\
\texttt{kimi-k2.6} & $88.49 \pm 0.15$ & $88 \pm 9$ \\
\texttt{gpt-5.6-terra} & $86.66 \pm 1.20$ & $28 \pm 2$ \\
\texttt{gpt-5.6-sol} & $86.40 \pm 2.53$ & $62 \pm 11$ \\
\texttt{muse-spark-1.1} & $83.19 \pm 12.03$ & $75 \pm 3$ \\
\texttt{glm-5.2} & $83.17 \pm 0.92$ & $58 \pm 3$ \\
\texttt{grok-4.5} & $81.82 \pm 9.87$ & $167 \pm 47$ \\
\texttt{qwen3.7-max} & $80.66 \pm 10.38$ & $54 \pm 5$ \\
\texttt{deepseek-v4-pro} & $79.15 \pm 2.67$ & $194 \pm 69$ \\
\texttt{gemini-3-flash} & $79.06 \pm 13.45$ & $51 \pm 5$ \\
\texttt{claude-sonnet-5} & $75.75 \pm 5.03$ & $46 \pm 6$ \\
\texttt{claude-opus-4.8} & $75.02 \pm 2.46$ & $30 \pm 3$ \\
\texttt{gemini-3.5-flash} & $74.59 \pm 11.02$ & $160 \pm 12$ \\
\texttt{minimax-m3} & $68.37 \pm 12.33$ & $74 \pm 9$ \\
\texttt{claude-haiku-4.5} & $36.90 \pm 22.73$ & $87 \pm 40$ \\
heuristic & $17.05 \pm 12.34$ & --- \\
idle & $-0.90 \pm 1.86$ & --- \\
random & $-17.21 \pm 2.45$ & --- \\
\bottomrule
\end{tabular}
\end{table}

\begin{figure}[t]
\centering
\includegraphics[width=0.75\textwidth]{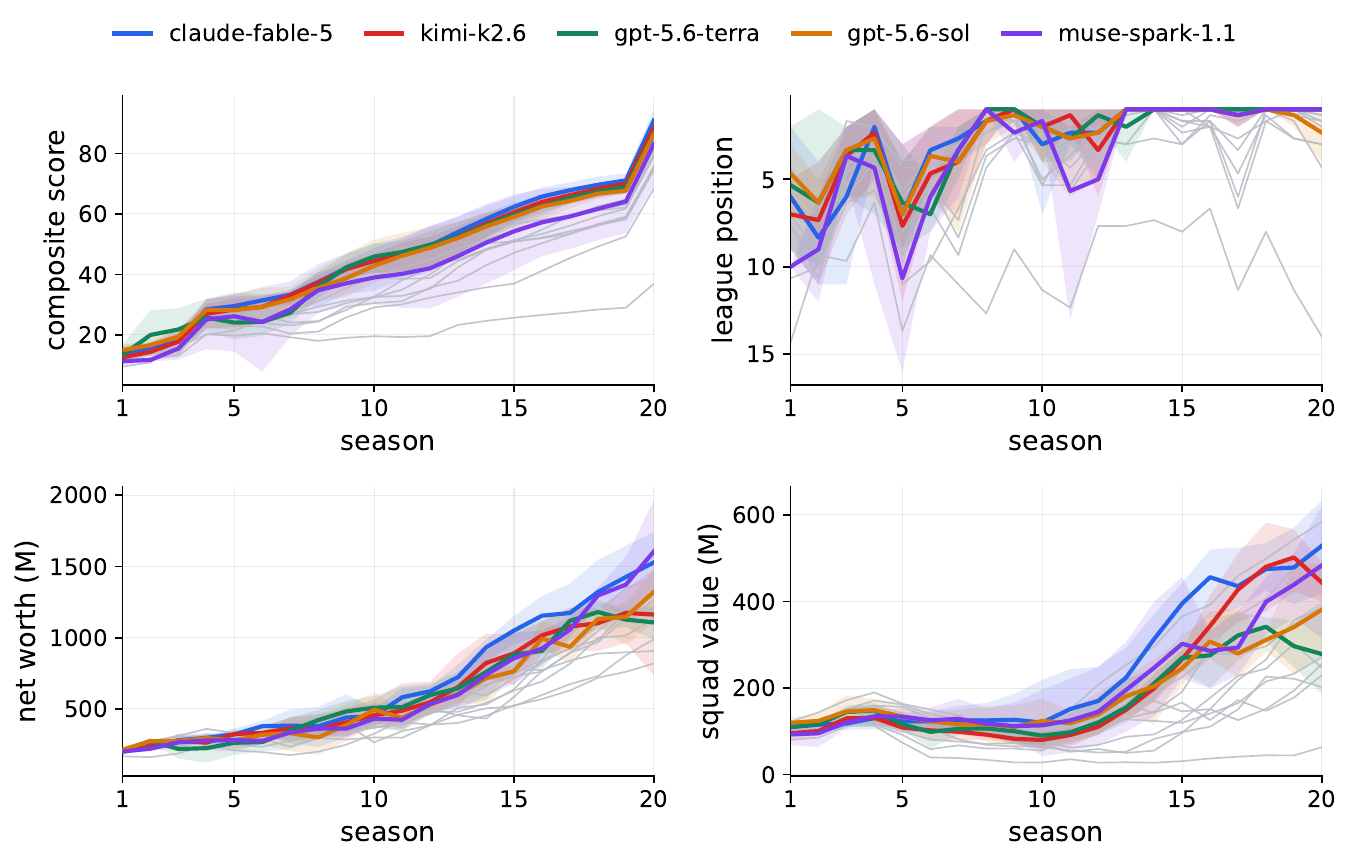}
\caption{Per-season solo trajectories for all 15 models in four channels. Lines are the mean over three seeds, shading spans the best and worst seed, and the five models with
the highest mean are colored and named in the legend; the grey lines
are the remaining ten models.}
\label{fig:solo-traj}
\end{figure}

\subsection{The 20-year solo track}\label{sec:campaign}

The solo track runs one tested model against 15 scripted clubs. An equal-endowment draft builds every squad from the same 50-template pool on the same budget, conflicting transfer offers resolve as first-price sealed bids with undisclosed clearing prices, and capped revival converts firing or insolvency into at most three restarts, discounted by $0.8^k$ after $k$ deaths, so a run reports full-horizon ability rather than a few unlucky seasons while the second chances never pay. The 15 opponents are the same scripted manager at five \emph{easy}, five \emph{medium}, and five \emph{hard} competence settings, differing in skill and never in information, so every model faces the same world (Appendix~\ref{app:tiers}). We ran three seeds for the 15 models (Appendix~\ref{app:roster}) and for three scripted anchors, including a disciplined heuristic, an idle policy, and a random policy (Appendix~\ref{app:anchors}).

\begin{figure}[t]
\centering
\includegraphics[width=0.8\textwidth]{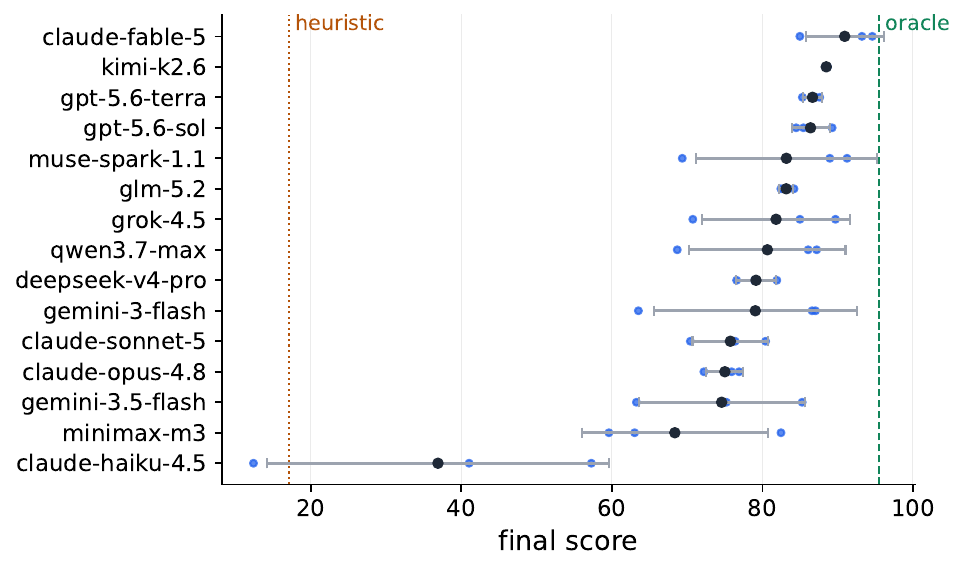}
\caption{Final score per model, mean with standard deviation over the
three seeds, individual seeds shown as points. Dashed and dotted lines
mark the oracle and heuristic means. Stability separates models that a
single-seed board cannot tell apart, from \texttt{kimi-k2.6} at 0.15 to
\texttt{claude-haiku-4.5} at 22.73.}
\label{fig:solo-stability}
\end{figure}

The oracle, a \emph{scripted} policy whose sole privilege is reading the true hidden state through the same interface and budgets, averages 95.54, and \texttt{claude-fable-5} reaches 90.94 without seeing any of it. Both sit at roughly two-thirds of a loose upper bound of ${\approx}145$ computed from channel values no run can reach (Appendix~\ref{app:score}), so headroom remains, and the scale is not saturated. Per-model spread across seeds runs from 0.15 for \texttt{kimi-k2.6} to 22.73 for \texttt{claude-haiku-4.5} (Figure~\ref{fig:solo-stability}), and four models swing by more than 20 points, so a strong mean can hide a collapse. Neither scale, price, nor vendor orders the ranking, as \texttt{gpt-5.6-terra} (86.66) finishes above \texttt{gpt-5.6-sol} (86.40), open-weight \texttt{kimi-k2.6} (88.49) clears both, and token spend is uncorrelated under any accounting (\S\ref{sec:capabilities}). Early standing does not predict the outcome either. On seed~1 the best-to-worst gap widens from 17.2 points at year~5 to 37.3 at year~20, the rank correlation with the final order climbs from 0.19 at year~5 to 0.78 at year~15, and \texttt{deepseek-v4-pro} leads on score at both year~5 and year~10 yet finishes 12th, while the eventual winner \texttt{claude-fable-5} is 5th at year~5. A shorter horizon would have ranked a different set of models. Figure~\ref{fig:solo-traj} plots every model season by season, and Table~\ref{tab:snapshots-solo} gives the numeric snapshots.

\begin{table}[t]
\centering
\small
\caption{The six first-play human runs (\S\ref{sec:human}).
Deaths are capped revivals spent before the settling firing; $t$ is the
game-year at which a fired run settled.}
\label{tab:human}
\begin{tabular}{@{}lrlrrr@{}}
\toprule
player & $S_{\text{final}}$ & settle & deaths & stops & time \\
\midrule
H1 & 74.64 & completed & 0 & 394 & 10\,h \\
H2 & 59.95 & completed & 0 & 398 & 4\,h \\
H3 & 5.39 & fired, $t{=}10.8$ & 4 & 202 & 4\,h \\
H4 & 2.47 & fired, $t{=}10.7$ & 4 & 198 & 5\,h \\
H5 & 0.41 & fired, $t{=}7.8$ & 4 & 165 & 1.5\,h \\
H6 & $-3.78$ & fired, $t{=}12.3$ & 4 & 237 & 2\,h \\
\bottomrule
\end{tabular}
\end{table}

\subsubsection{Capability decomposition: what the ranking is made of}\label{sec:capabilities}

A composite score orders the field but hides why a model lands where it does, and evaluation should separate outcome from the behavior that produced it. We decompose solo performance into six behavioral capabilities (Figure~\ref{fig:capmatrix}), each read from a reproducible behavioral metric whose exact construction is given in Appendix~\ref{app:behavior}. Associations with the final score are Spearman rank correlations over the 15 models, written $r_s$.

\textbf{Endgame awareness.} A 20-year run ends, and slow-payoff actions stop being investments once there is no time left to collect. We measure the late-minus-mid shift in facility and academy actions, the rate over years 17--20 minus the rate over years 2--16, and it is the capability that tracks the score most closely ($r_s = -0.58$, same sign on every seed: $-0.31$ / $-0.34$ / $-0.28$). The winner cuts hardest, from 2.4 to 0.8 such actions per season, \texttt{gpt-5.6-terra} from 1.4 to 0.4, while \texttt{gemini-3.5-flash} instead \emph{raises} them from 2.4 to 3.3 and is still building facilities in year 19. Rational endgame behavior under a finite horizon is a question only a long benchmark can pose; in short episodes it does not exist.

\begin{figure}[t]
\centering
\includegraphics[width=0.85\textwidth]{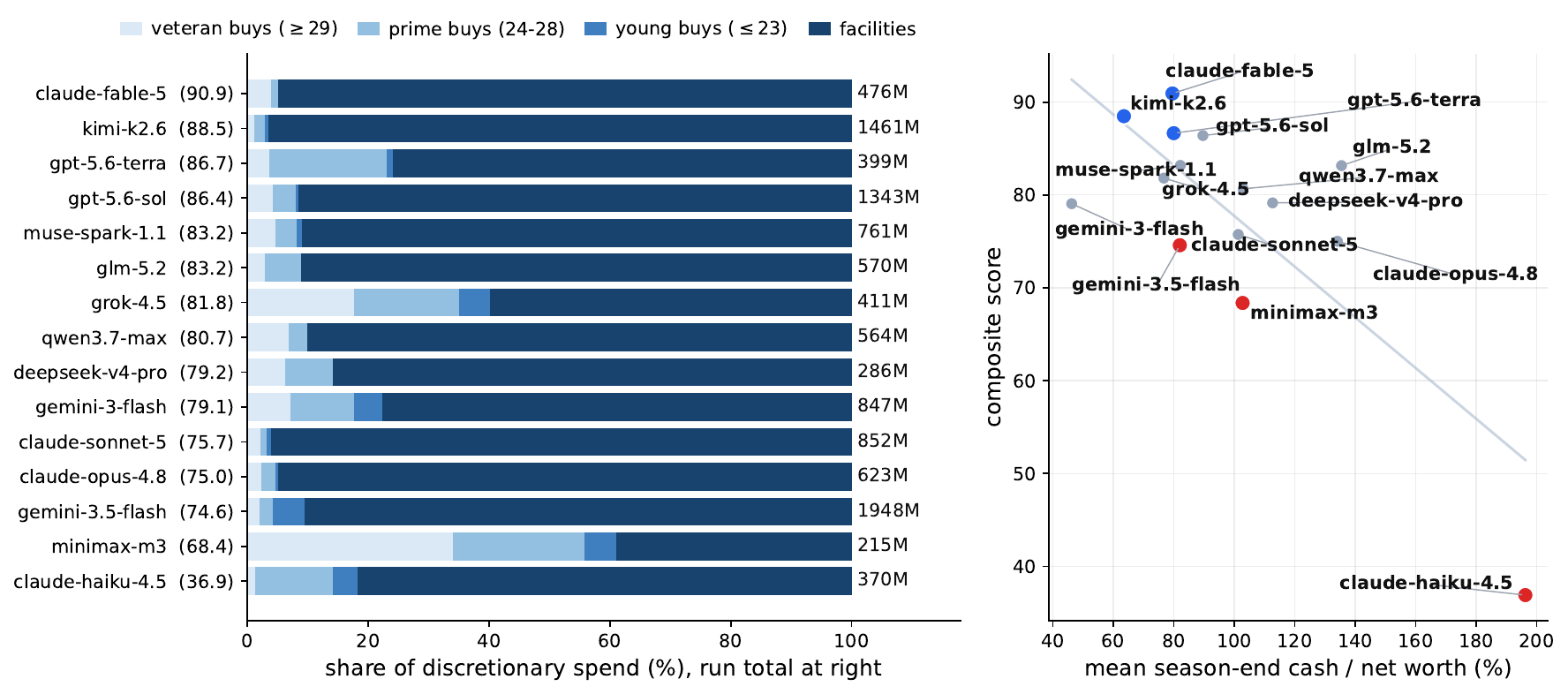}
\caption{Credit assignment. Left: discretionary spend bucketed by payoff
horizon, with run totals at
right. Right: mean season-end idle-cash ratio vs.\ final score,
$r_s = -0.50$, negative on every seed.}
\label{fig:idlecash}
\end{figure}

\textbf{Credit assignment.} The mean ratio of season-end cash to net worth tracks the score across all three seeds ($r_s = -0.50$, and $-0.74$ / $-0.60$ / $-0.34$ per seed; Figure~\ref{fig:idlecash}). Net worth books excess cash at a discount, so a ratio above 100\% means the idle pile exceeds the club's entire discounted worth, and the bottom of the board sits there: \texttt{claude-haiku-4.5} averages 196\%, \texttt{glm-5.2} 135\%, \texttt{claude-opus-4.8} 134\%, against 80\% for the winner and a field median of 90\%. Total facility investment is uncorrelated, so the signal is \emph{whether money sits idle}, not whether it is spent. Holding cash is the locally safe action whose opportunity cost surfaces seasons later, exactly the credit horizon the environment stresses (demand 2).

\begin{figure}[ht]
\centering
\includegraphics[width=0.85\textwidth]{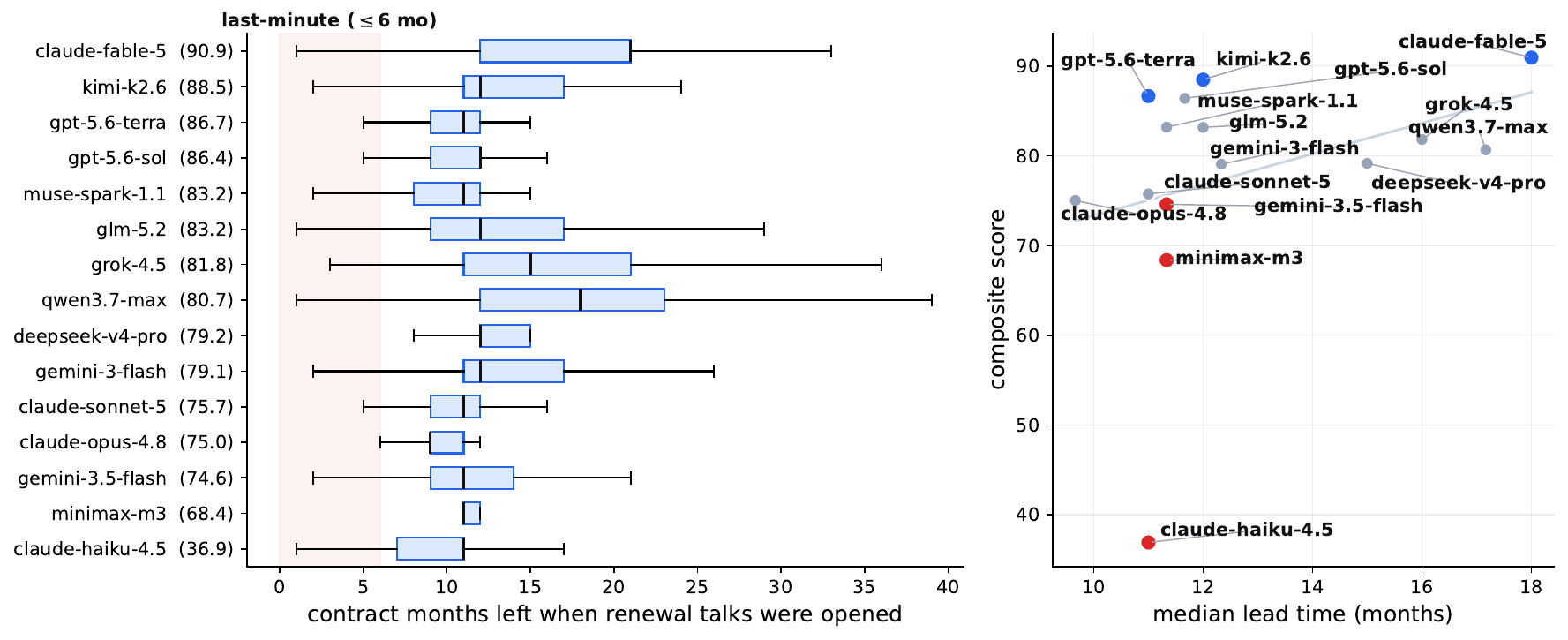}
\caption{Proactive control. Left: distribution of contract months remaining
when each renewal negotiation episode opens (rows sorted by final score;
shaded band = last-minute, $\leq 6$ months). Right: per-model median lead
vs.\ final score, $r_s = +0.45$, positive on every seed.}
\label{fig:renewal}
\end{figure}

\textbf{Proactive control.} For a fully foreseeable future event, a contract expiring, the field splits by \emph{when} it acts (Figure~\ref{fig:renewal}). The winner opens renewal negotiations a median of 18 months before expiry, with 4\% of episodes opened inside the final six months, while \texttt{claude-opus-4.8} opens at a median of 10 months with 20\% last-minute and \texttt{claude-haiku-4.5} at 11 months with 21\% ($r_s = +0.45$, positive on every seed: $+0.68$ / $+0.18$ / $+0.47$). The early-renewal policy is not unique to the winner. \texttt{qwen3.7-max} holds the longest renewal leads after the winner, and one first-play human derived the same policy, signing every contract for the maximum length (\S\ref{sec:human}).

\textbf{Price discovery.} The oracle bids the seller's true accept threshold and closes every buy on the first offer (Appendix~\ref{app:oracle}), fixing the reference at 1.0 offers per completed signing. No model comes close: the field median is 30, the best model (\texttt{claude-fable-5}) needs 9, and \texttt{gemini-3.5-flash} needs 73, with single seeds as high as 133. After twenty years and hundreds of rejections, the models have not learned where the acceptance boundary lies; the low completion rate of the solo market is a pricing failure, not an empty market. The behavior is a judgment about the market, not a fixed habit: the winner bid 0.5 times per season in the solo market but 4.6 in the Arena (\S\ref{sec:arena}).

\begin{figure}[t]
\centering
\includegraphics[width=\textwidth]{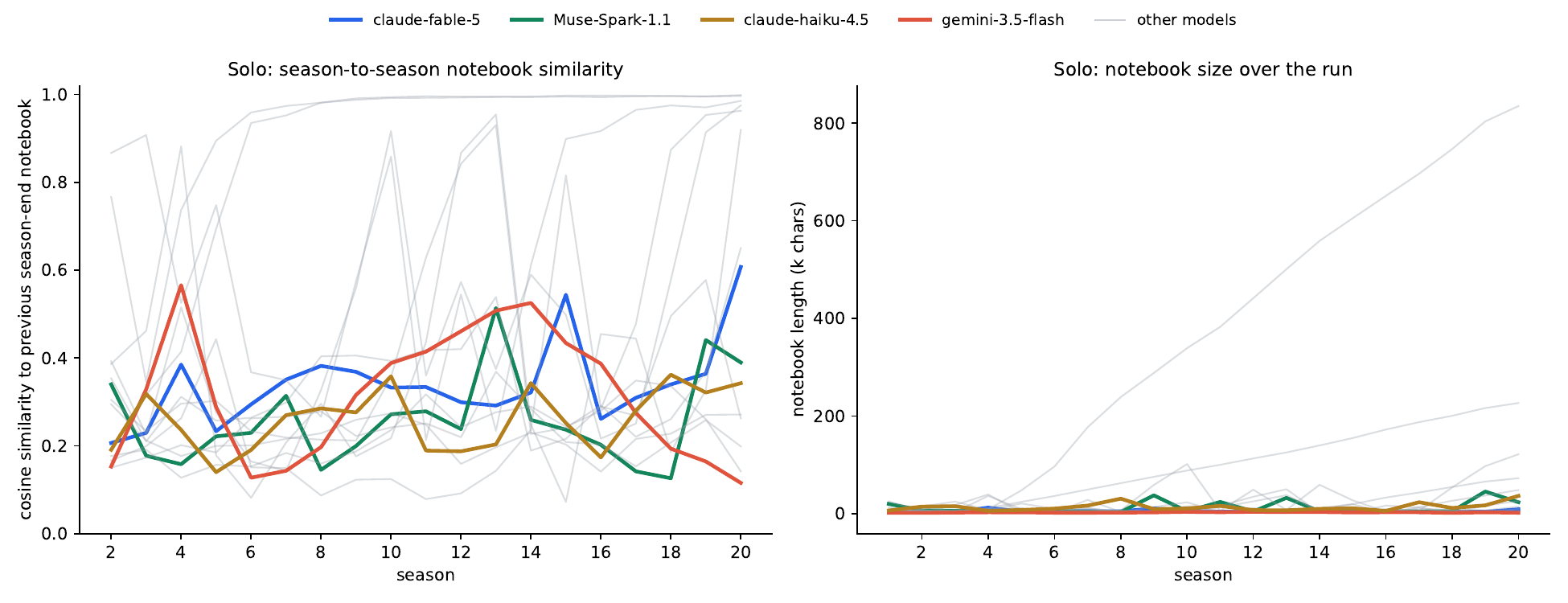}
\caption{Memory-curation regimes on the solo track. Left: season-to-season
TF--IDF cosine similarity of each model's reconstructed season-end notebook.
Right: notebook size over the run. The right panel decodes the left: the
same ``high consistency'' is a 200k-character append-only archive for one
model and a 3--6k curated document for the winner.}
\label{fig:memcurves}
\end{figure}

\textbf{Memory curation.} The notebook is the only cross-stop state the agent controls (\S\ref{sec:memory}). We reconstruct each model's notebook at every season end and take the TF--IDF cosine similarity between consecutive snapshots, so 1 means the text never changes, and 0 means a full rewrite. This similarity separates two opposite failures (Figure~\ref{fig:memcurves}, Appendix~\ref{app:behavior}): \texttt{gpt-5.6-sol} sits at 0.91, an append-only archive in which current state drowns in history, while \texttt{claude-sonnet-5} (0.20) and \texttt{qwen3.7-max} (0.23) rewrite so completely that no plan survives to be executed. The winner sits at 0.39 against a field median of 0.31, holding a stable strategy skeleton while rewriting the state each season. Similarity alone does not certify good curation, though. \texttt{claude-haiku-4.5} averages 0.31, as close to the middle band as the winner, and still finishes last.

\begin{figure}[t]
\centering
\includegraphics[width=0.62\textwidth]{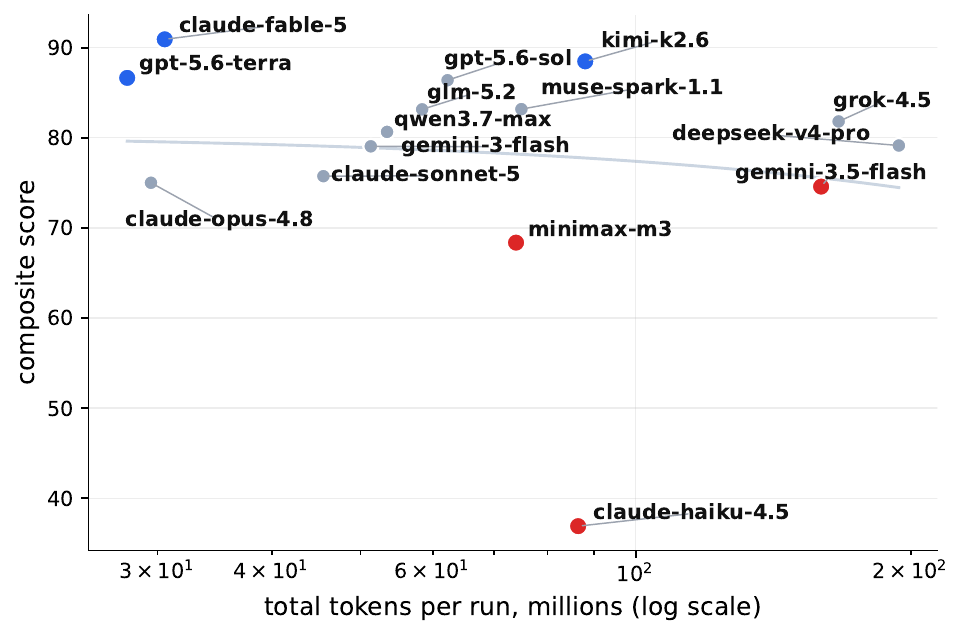}
\caption{Compute allocation. Mean tokens per run against mean solo score,
both over the three seeds, on a log axis. A null result
($r_s = -0.19$, $p = 0.50$) across a sevenfold spend range.}
\label{fig:tokens}
\end{figure}

\textbf{Compute allocation and efficiency.} Token spend spans a factor of seven, from 28M for \texttt{gpt-5.6-terra} to 194M for \texttt{deepseek-v4-pro}, yet it does not order the board on any seed ($r_s = -0.19$, and $+0.07$ / $-0.23$ / $-0.15$ per seed; Figure~\ref{fig:tokens}), and the winner is among the three cheapest models. The matrix's compute axis (Figure~\ref{fig:capmatrix}) therefore reports efficiency, score per million tokens; \texttt{gpt-5.6-terra} and the winner top it, and the heaviest spenders fill its bottom.

\begin{figure}[t]
\centering
\includegraphics[width=0.66\textwidth]{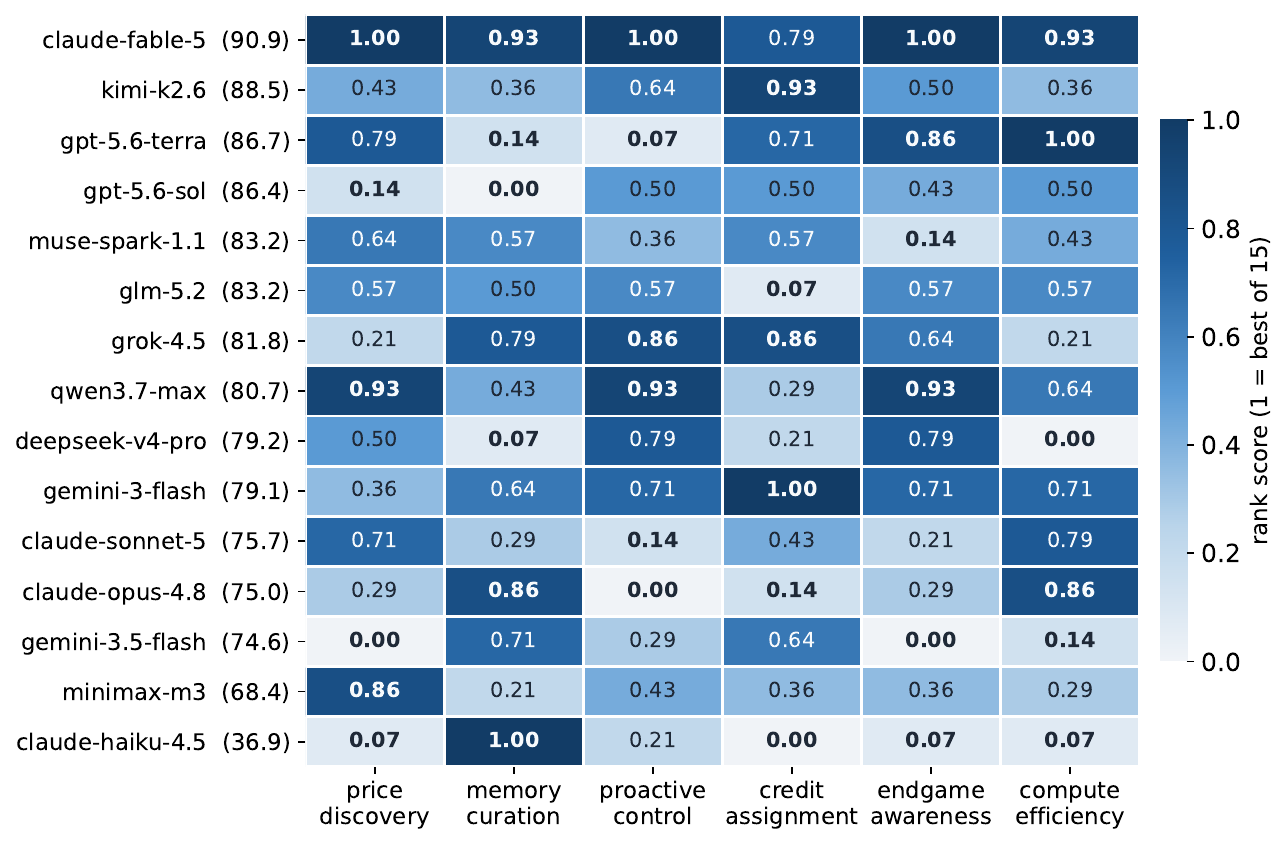}
\caption{Capability matrix over the 15 solo models, six axes. Each column
rank-normalizes one behavioral metric averaged over the three seeds
(1 = best of 15; construction in Appendix~\ref{app:behavior}), and rows are
sorted by mean final score. The winner is a generalist, with no axis below 0.79
and mean 0.94, rather than the leader of any single one; the mid-table pairs
a genuine strength with a decisive gap; and the bottom of the board is weak
on most axes at once.}
\label{fig:capmatrix}
\end{figure}

\textbf{Capability profiles.} The winner is even across the axes, topping the board without leading most single ones (Figure~\ref{fig:capmatrix}; per-model profiles in Table~\ref{tab:profiles}, Appendix~\ref{app:behavior}). The mid-table agents pair one genuine strength with one clear gap. \texttt{gemini-3-flash} keeps the best cash discipline in the field (46\% idle-cash ratio against a median of 90\%) yet ranks tenth, the model least willing to reduce long-horizon spending late in the run. The bottom fails on several axes at once, \texttt{claude-haiku-4.5} pairing the highest idle cash with the shortest renewal leads and a rising endgame investment rate.

\subsubsection{Human study}\label{sec:human}

Six people played the same 20-year track in web-play mode, none of whom had played FM-Bench before. Four of the six died out (Table~\ref{tab:human}), all four below the disciplined heuristic anchor. The other two completed the horizon with zero deaths, at 74.64 and 59.95, clear of every scripted anchor. Against the three-seed model means of Table~\ref{tab:campaign}, the weaker survivor finishes above only \texttt{claude-haiku-4.5} (36.90) and the stronger lands level with \texttt{gemini-3.5-flash} (74.59), 16 points short of \texttt{claude-fable-5}, placing frontier models between untrained humans and the oracle. Asked what they found hard (Appendix~\ref{app:humanfeedback}), the players named the demands, from unrecoverable bids against an unseen accept threshold to a wage bill one bad season from dismissal. Their failures mirror the decomposition, letting cash idle while the bill arrived seasons later (\S\ref{sec:capabilities}), while their strengths are the ones the models lack, deriving standing rules, revising failing policies mid-run, and building an external helper for implicit state. The two profiles are close to complementary, and since the score is identical across modes, whether a human--agent pairing beats either alone, and which capability the gain comes from, is measurable. We leave that measurement to future work.

\begin{table}[t]
\centering
\small
\caption{The Arena results.}
\label{tab:arena}
\begin{tabular}{rlrrlr}
\toprule
\# & seat & $S_{\text{final}}$ & deaths & settle & tokens \\
\midrule
1 & claude-fable-5 & 76.26 & 0 & completed & ${\sim}$34M \\
2 & muse-spark-1.1 & 62.47 & 0 & completed & ${\sim}$80M \\
3 & deepseek-v4-pro & 52.09 & 0 & completed & ${\sim}$111M \\
4 & glm-5.2 & 51.44 & 0 & completed & ${\sim}$86M \\
5 & grok-4.5 & 50.89 & 0 & completed & ${\sim}$138M \\
6 & gpt-5.6-sol & 47.94 & 0 & completed & ${\sim}$78M \\
7 & minimax-m3 & 44.78 & 0 & completed & ${\sim}$65M \\
8 & claude-sonnet-5 & 40.75 & 0 & completed & ${\sim}$42M \\
9 & kimi-k2.6 & 39.72 & 0 & completed & ${\sim}$119M \\
10 & gpt-5.6-terra & 38.44 & 0 & completed & ${\sim}$19M \\
11 & qwen3.7-max & 32.77 & 2 & completed & ${\sim}$54M \\
12 & claude-opus-4.8 & 28.44 & 1 & completed & ${\sim}$43M \\
13 & gemini-3-flash & 15.76 & 2 & completed & ${\sim}$63M \\
14 & claude-haiku-4.5 & 0.76 & 4 & fired, $t{=}10.7$ & ${\sim}$30M \\
15 & gemini-3.5-flash & 0.21 & 4 & fired, $t{=}5.8$ & ${\sim}$40M \\
16 & heuristic (anchor) & 0.13 & 4 & fired, $t{=}2.6$ & --- \\
\bottomrule
\end{tabular}
\end{table}
\subsection{The Arena: a shared-world league}\label{sec:arena}

\begin{figure}[t]
\centering
\includegraphics[width=0.75\linewidth]{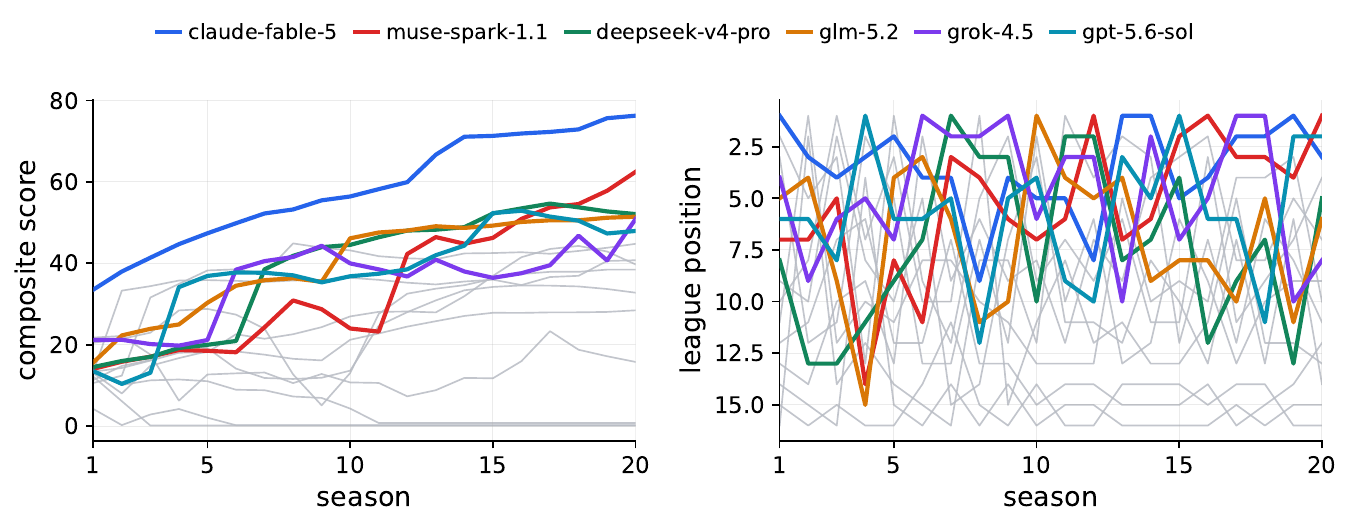}
\caption{Arena trajectories, all 16 seats. The six highest-scoring
seats are colored and named in the legend; the grey lines are the
remaining ten seats. Left: composite $S_{\text{final}}$ at each season
end. Right: league-table position.}
\label{fig:arena-traj}
\end{figure}

The Arena places 15 agents and the heuristic anchor in one shared 20-year world under the same stack as the solo track (\S\ref{sec:campaign}). Table~\ref{tab:arena} reports all 16 seats. The per-season trajectories (Figure~\ref{fig:arena-traj}, asset channels in Figure~\ref{fig:arena-assets}, Appendix~\ref{app:behavior}) again put the scripted anchor first out of the world, in year 3, but the shared world is harsher than the solo track, and the two weakest LLM models follow it out by exhausting their revivals. They also show league standings and composite scores diverging by design, since \texttt{gemini-3-flash} led the table at year 5 yet finished 13th after two late firings, while the eventual winner spent the mid-game in mid-table accumulating convertible assets. As on the solo track, scale and price do not order the board, with several low-cost open-weight models above flagship-priced ones.

\paragraph{Dynasties on the solo track, rotation in the Arena.} On the solo track, the four highest-scoring runs each hold the league title continuously through year 20 (Figure~\ref{fig:solo-traj}). In the Arena, the title rotates. Ten different models won it at least once, the reigning champion kept it in only 2 of 19 season transitions, and the eventual composite winner \texttt{claude-fable-5} took just 4 titles. The contrast isolates the counter-adaptive market (demand 3). Against fixed opponents, an early lead compounds into a rich-get-richer dynasty, while adaptive rivals bid talent away from whoever leads, and the composite winner is set by 20-year asset accumulation rather than title count. The winner feels the same pressure in its own trading, 0.5 transfer offers per season against the fixed solo opponents but 4.6 in the contested Arena market.

\paragraph{Behavioral analysis.} The ordering summarizes outcomes; to see the behavior behind it, we read the full action logs and self-authored notebooks of three models, the winner (\texttt{claude-fable-5}), the early leader that collapsed (\texttt{gemini-3-flash}), and the highest-cash low-placing model (\texttt{claude-opus-4.8}).

\emph{(i) Persistent planning versus season-local reaction.} The winner kept one cumulative plan for all 20 years, a running honors ledger whose rewrites revise a single strategy rather than starting over, and intervened sparingly and precisely, from a Y13 wage-breach remedy to a Y19 title push, on 91 transfer offers in total. \texttt{gemini-3-flash} issued 452 offers and 2{,}642 total actions (the winner's 1{,}315), and its notebook resets to season-local firefighting at each crisis. Table position bought by heavy early activity did not survive the horizon.

\emph{(ii) A knowing--doing gap on delayed reward.} \texttt{claude-opus-4.8} recorded the correct policy in writing, noting at Y10 that idle cash should be deployed on quality and at Y19 that its reserves were discounted and should become young players, yet finished holding some 2{,}100M in idle reserves. Holding cash is locally safe but earns almost nothing under the composite. The model wrote the right long-range plan and did not execute it.

\emph{(iii) Deliberation volume does not predict skill.} The compute null of \S\ref{sec:capabilities} holds under every accounting we tried, cache-inclusive, output-only, and dollars (all $p > 0.38$, $n = 15$), and since each model chooses its own number of turns per stop, the spread is a behavioral property rather than a harness setting.

\section{Conclusion}\label{sec:conclusion}

FM-Bench measures whether language model agents can manage rather than merely execute, handing each one a football club for twenty in-game years under the four demands that define the job, with a deterministic engine grading every run and no LLM judge. Every frontier model completes the horizon that the blind scripts do not survive, and the strongest, \texttt{claude-fable-5}, tops both tracks while reaching about 95\% of a privileged oracle, so the benchmark separates models without being saturated. The board is one no shorter or cheaper measurement would predict. Scale, price, and vendor do not order it, year-5 standings barely correlate with the final order, and the title rotates among ten models once the rivals are adaptive. The scalar decomposes into interpretable behaviors, and the claim that one model manages better unpacks into which demands it meets, which it fails to meet, and what it was doing when it failed.

\clearpage
\bibliographystyle{plainnat}
\bibliography{references}

@inproceedings{liu2024agentbench,
  title={Agentbench: Evaluating llms as agents},
  author={Liu, Xiao and Yu, Hao and Zhang, Hanchen and Xu, Yifan and Lei, Xuanyu and Lai, Hanyu and Gu, Yu and Ding, Hangliang and Men, Kaiwen and Yang, Kejuan and others},
  booktitle={International Conference on Learning Representations},
  volume={2024},
  pages={52989--53046},
  year={2024}
}

@inproceedings{qin2024toolllm,
  title={Toolllm: Facilitating large language models to master 16000+ real-world apis},
  author={Qin, Yujia and Liang, Shihao and Ye, Yining and Zhu, Kunlun and Yan, Lan and Lu, Yaxi and Lin, Yankai and Cong, Xin and Tang, Xiangru and Qian, Bill and others},
  booktitle={International Conference on Learning Representations},
  volume={2024},
  pages={9695--9717},
  year={2024}
}

@article{yao2024tau,
  title={$\tau$-bench: A Benchmark for Tool-Agent-User Interaction in Real-World Domains},
  author={Yao, Shunyu and Shinn, Noah and Razavi, Pedram and Narasimhan, Karthik},
  journal={arXiv preprint arXiv:2406.12045},
  year={2024}
}

@inproceedings{zhou2024webarena,
  title={Webarena: A realistic web environment for building autonomous agents},
  author={Zhou, Shuyan and Xu, Frank F and Zhu, Hao and Zhou, Xuhui and Lo, Robert and Sridhar, Abishek and Cheng, Xianyi and Ou, Tianyue and Bisk, Yonatan and Fried, Daniel and others},
  booktitle={International Conference on Learning Representations},
  volume={2024},
  pages={15585--15606},
  year={2024}
}

@article{xie2024osworld,
  title={Osworld: Benchmarking multimodal agents for open-ended tasks in real computer environments},
  author={Xie, Tianbao and Zhang, Danyang and Chen, Jixuan and Li, Xiaochuan and Zhao, Siheng and Cao, Ruisheng and Hua, Toh J and Cheng, Zhoujun and Shin, Dongchan and Lei, Fangyu and others},
  journal={Advances in Neural Information Processing Systems},
  volume={37},
  pages={52040--52094},
  year={2024}
}

@article{kuttler2020nethack,
  title={The nethack learning environment},
  author={K{\"u}ttler, Heinrich and Nardelli, Nantas and Miller, Alexander and Raileanu, Roberta and Selvatici, Marco and Grefenstette, Edward and Rockt{\"a}schel, Tim},
  journal={Advances in Neural Information Processing Systems},
  volume={33},
  pages={7671--7684},
  year={2020}
}

@article{hafner2021benchmarking,
  title={Benchmarking the spectrum of agent capabilities},
  author={Hafner, Danijar},
  journal={arXiv preprint arXiv:2109.06780},
  year={2021}
}

@inproceedings{cote2018textworld,
  title={Textworld: A learning environment for text-based games},
  author={C{\^o}t{\'e}, Marc-Alexandre and K{\'a}d{\'a}r, Akos and Yuan, Xingdi and Kybartas, Ben and Barnes, Tavian and Fine, Emery and Moore, James and Hausknecht, Matthew and El Asri, Layla and Adada, Mahmoud and others},
  booktitle={Workshop on computer games},
  pages={41--75},
  year={2018},
  organization={Springer}
}

@inproceedings{paglieri2025balrog,
  title={Balrog: Benchmarking agentic llm and vlm reasoning on games},
  author={Paglieri, Davide and Cupia{\l}, Bart{\l}omiej and Coward, Samuel and Piterbarg, Ulyana and Wo{\l}czyk, Maciej and Khan, Akbir and Pignatelli, Eduardo and Kuci{\'n}ski, {\L}ukasz and Pinto, Lerrel and Fergus, Rob and others},
  booktitle={International Conference on Learning Representations},
  volume={2025},
  pages={96666--96702},
  year={2025}
}

@article{backlund2025vendingbench,
  title={Vending-bench: A benchmark for long-term coherence of autonomous agents},
  author={Backlund, Axel and Petersson, Lukas},
  journal={arXiv preprint arXiv:2502.15840},
  year={2025}
}

@inproceedings{jimenez2024swe,
  title={Swe-bench: Can language models resolve real-world github issues?},
  author={Jimenez, Carlos E and Yang, John and Wettig, Alexander and Yao, Shunyu and Pei, Kexin and Press, Ofir and Narasimhan, Karthik},
  booktitle={International Conference on Learning Representations},
  volume={2024},
  pages={54107--54157},
  year={2024}
}

@inproceedings{mialon2024gaia,
  title={Gaia: a benchmark for general ai assistants},
  author={Mialon, Gr{\'e}goire and Fourrier, Cl{\'e}mentine and Wolf, Thomas and LeCun, Yann and Scialom, Thomas},
  booktitle={International Conference on Learning Representations},
  volume={2024},
  pages={9025--9049},
  year={2024}
}

@article{coffeebench2026,
  title={CoffeeBench: Benchmarking Long-Horizon LLM Agents in Heterogeneous Multi-Agent Economies},
  author={Sugiura, Issa and Hattori, Daichi and Araragi, Kazuo and Ogawa, Keita and Onose, Shota and Makino, Taro and Usuki, Teppei and Ishida, Takashi},
  journal={arXiv preprint arXiv:2606.16613},
  year={2026}
}

@article{agentatlas2026,
  title={AgentAtlas: Beyond Outcome Leaderboards for LLM Agents},
  author={Mazaheri, Parsa and Mazaheri, Kasra},
  journal={arXiv preprint arXiv:2605.20530},
  year={2026}
}

@article{ultrahorizon2025,
  title={Ultrahorizon: Benchmarking agent capabilities in ultra long-horizon scenarios},
  author={Luo, Haotian and Zhang, Huaisong and Zhang, Xuelin and Wang, Haoyu and Qin, Zeyu and Lu, Wenjie and Ma, Guozheng and He, Haiying and Xie, Yingsha and Zhou, Qiyang and others},
  journal={arXiv preprint arXiv:2509.21766},
  year={2025}
}

@inproceedings{lightman2024let,
  title={Let's verify step by step},
  author={Lightman, Hunter and Kosaraju, Vineet and Burda, Yuri and Edwards, Harrison and Baker, Bowen and Lee, Teddy and Leike, Jan and Schulman, John and Sutskever, Ilya and Cobbe, Karl},
  booktitle={International Conference on Learning Representations},
  volume={2024},
  pages={39578--39601},
  year={2024}
}

@article{balunovic2026matharena,
  title={Matharena: Evaluating llms on uncontaminated math competitions},
  author={Balunovic, Mislav and Dekoninck, Jasper and Petrov, Ivo and Jovanovi{\'c}, Nikola and Vechev, Martin},
  journal={Advances in Neural Information Processing Systems},
  volume={38},
  year={2026}
}

@article{cheng2026terminal,
  title={Terminal-world: Scaling terminal-agent environments via agent skills},
  author={Cheng, Zihao and Wang, Hongru and Liu, Zeming and Wang, Xinyi and Zhu, Xiangrong and Guo, Yuhang and Lin, Wei and Pan, Jeff Z and Wang, Yunhong},
  journal={arXiv preprint arXiv:2605.20876},
  year={2026}
}

@article{shi2026merchantbench,
  title={MerchantBench: Benchmarking LLM Agents for Long-Term Coherence in E-Commerce Operations},
  author={Shi, Qiming and Tao, Yulong and Jin, Linbo and Kang, Zhaolu and Dou, Yibo and Zhu, Jiawen and Pan, Tianjun and Fu, Shaokang and Wang, Chengyu and Li, Siyue and others},
  journal={arXiv preprint arXiv:2607.28956},
  year={2026}
}

@article{muller2026cattle,
  title={Cattle Trade: A Multi-Agent Benchmark for LLM Bluffing, Bidding, and Bargaining},
  author={M{\"u}ller, Robert and M{\"u}ller, Clemens},
  journal={arXiv preprint arXiv:2605.14537},
  year={2026}
}

@article{li2026long,
  title={Long-Horizon-Terminal-Bench: Testing the Limits of Agents on Long-Horizon Terminal Tasks with Dense Reward-Based Grading},
  author={Li, Zongxia and Li, Zhongzhi and Shi, Yucheng and Wang, Ruhan and Yang, Junyao and Liu, Zhichao and Wu, Xiyang and Li, Anhao and Yu, Yue and Liu, Ninghao and others},
  journal={arXiv preprint arXiv:2607.08964},
  year={2026}
}

@article{wang2024battleagentbench,
  title={Battleagentbench: A benchmark for evaluating cooperation and competition capabilities of language models in multi-agent systems},
  author={Wang, Wei and Zhang, Dan and Feng, Tao and Wang, Boyan and Tang, Jie},
  journal={arXiv preprint arXiv:2408.15971},
  year={2024}
}

@inproceedings{zhu2025multiagentbench,
  title={Multiagentbench: Evaluating the collaboration and competition of llm agents},
  author={Zhu, Kunlun and Du, Hongyi and Hong, Zhaochen and Yang, Xiaocheng and Guo, Shuyi and Wang, Daisy Zhe and Wang, Zhenhailong and Qian, Cheng and Tang, Xiangru and Ji, Heng and others},
  booktitle={Proceedings of the 63rd Annual Meeting of the Association for Computational Linguistics (Volume 1: Long Papers)},
  pages={8580--8622},
  year={2025}
}

@article{cfe,
  title={Classroom Final Exam: An Instructor-Tested Reasoning Benchmark},
  author={Gao, Chongyang and Yang, Diji and Zhou, Shuyan and Yan, Xichen and Song, Luchuan and Li, Shuo and Chen, Kezhen},
  journal={arXiv preprint arXiv:2602.19517},
  year={2026}
}

@inproceedings{he2026yc,
  title={YC-Bench: Benchmarking AI agents for long-term planning and consistent execution},
  author={He, Muyu and Tu, Vincent and Jain, Adit and Kumar, Anand and Patro, Sachin and Bakshi, Soumyadeep and Rajani, Nazneen},
  booktitle={ICML 2026 Workshop on Combining Theory and Benchmarks: Towards A Virtuous Cycle to Understand and Guarantee Foundation Model Performance},
  year={2026}
}

@article{chen2025stockbench,
  title={Stockbench: Can llm agents trade stocks profitably in real-world markets?},
  author={Chen, Yanxu and Yao, Zijun and Liu, Yantao and Xin, Amy and Ye, Jin and Yu, Jianing and Hou, Lei and Li, Juanzi},
  journal={arXiv preprint arXiv:2510.02209},
  year={2025}
}

@article{cfo,
  title={Can LLM Agents Be CFOs? Benchmarking Long-Horizon Resource Allocation in an Uncertain Enterprise Environment},
  author={Han, Yi and Wang, Yan and Qian, Lingfei and Li, Haohang and Cao, Yupeng and He, Yueru and Peng, Xueqing and Shen, Nanhan and Xu, Yitao and Chen, Yankai and others},
  journal={arXiv preprint arXiv:2603.23638},
  year={2026}
}

@article{ceo,
  title={CEO-Bench: Can Agents Play the Long Game?},
  author={Chen, Haozhe and Narasimhan, Karthik and Liu, Zhuang},
  journal={arXiv preprint arXiv:2606.18543},
  year={2026}
}

@misc{andonlabs2026arena,
  title={Vending-Bench Arena},
  author={{Andon Labs}},
  howpublished={\url{https://andonlabs.com/evals/vending-bench-arena}},
  year={2026},
  note={Leaderboard page and blog posts only; no technical report or source release}
}

@misc{anthropic2026fable5,
  author = {{Anthropic}},
  title  = {Claude Fable 5 and Claude Mythos 5},
  year   = {2026},
  note   = {System card and announcement},
  url    = {https://www.anthropic.com/news/claude-fable-5-mythos-5}
}

@misc{anthropic2026sonnet5,
  author = {{Anthropic}},
  title  = {Introducing Claude Sonnet 5},
  year   = {2026},
  url    = {https://www.anthropic.com/news/claude-sonnet-5}
}

@misc{anthropic2026opus48,
  author = {{Anthropic}},
  title  = {Claude Opus 4.8 System Card},
  year   = {2026},
  note   = {https://www.anthropic.com/news/claude-opus-4-8}
}

@misc{anthropic2025haiku45,
  author = {{Anthropic}},
  title  = {System Card: Claude Haiku 4.5},
  year   = {2025},
  url    = {https://www.anthropic.com/news/claude-haiku-4-5}
}

@misc{openai2026gpt56,
  author = {{OpenAI}},
  title  = {Previewing GPT-5.6 Sol, Terra and Luna},
  year   = {2026},
  note   = {System card at deploymentsafety.openai.com/gpt-5-6-preview},
  url    = {https://openai.com/index/previewing-gpt-5-6-sol/}
}

@misc{google2025gemini3flash,
  author = {{Google DeepMind}},
  title  = {Gemini 3 Flash Model Card},
  year   = {2025},
  url    = {https://deepmind.google/models/gemini/flash/}
}

@misc{google2026gemini35flash,
  author = {{Google DeepMind}},
  title  = {Gemini 3.5 Flash Model Card},
  year   = {2026},
  url    = {https://deepmind.google/models/model-cards/gemini-3-5-flash/}
}

@misc{xai2026grok45,
  author = {{xAI}},
  title  = {Grok 4.5},
  year   = {2026},
  note   = {https://x.ai/news/grok-4-5}
}

@misc{moonshot2026kimik26,
  author = {{Moonshot AI}},
  title  = {Kimi K2.6},
  year   = {2026},
  note   = {Model card},
  url    = {https://www.kimi.com/ai-models/kimi-k2-6}
}

@misc{alibaba2026qwen37max,
  author = {{Alibaba Qwen Team}},
  title  = {Qwen3.7-Max},
  year   = {2026},
  note   = {https://qwen.ai/blog?id=qwen3.7}
}

@misc{zhipu2026glm52,
  author = {{Zhipu AI}},
  title  = {GLM-5.2},
  year   = {2026},
  note   = {https://z.ai/blog/glm-5.2)}
}

@article{deepseek2026v4,
  title={Deepseek-v4: Towards highly efficient million-token context intelligence},
  author={Xu, Anyi and Lin, Bangcai and Xue, Bing and Wang, Bingxuan and Xu, Bingzheng and Wu, Bochao and Zhang, Bowei and Lin, Chaofan and Dong, Chen and Ling, Chenchen and others},
  journal={arXiv preprint arXiv:2606.19348},
  year={2026}
}

@misc{minimax2026m3,
  author = {{MiniMax Team}},
  title  = {MiniMax M3},
  year   = {2026},
  note   = {https://www.minimax.io/blog/minimax-m3}
}

@misc{meta2026musespark,
  author = {{Meta Superintelligence Labs}},
  title  = {Introducing Muse Spark 1.1},
  year   = {2026},
  url    = {https://ai.meta.com/blog/introducing-muse-spark-meta-model-api/}
}

@InProceedings{pmlr-v235-chiang24b,
  title = 	 {Chatbot Arena: An Open Platform for Evaluating {LLM}s by Human Preference},
  author =       {Chiang, Wei-Lin and Zheng, Lianmin and Sheng, Ying and Angelopoulos, Anastasios Nikolas and Li, Tianle and Li, Dacheng and Zhu, Banghua and Zhang, Hao and Jordan, Michael and Gonzalez, Joseph E. and Stoica, Ion},
  booktitle = 	 {Proceedings of the 41st International Conference on Machine Learning},
  pages = 	 {8359--8388},
  year = 	 {2024},
  editor = 	 {Salakhutdinov, Ruslan and Kolter, Zico and Heller, Katherine and Weller, Adrian and Oliver, Nuria and Scarlett, Jonathan and Berkenkamp, Felix},
  volume = 	 {235},
  series = 	 {Proceedings of Machine Learning Research},
  month = 	 {21--27 Jul},
  publisher =    {PMLR},
  url = 	 {https://proceedings.mlr.press/v235/chiang24b.html}
}

\beginappendix

\section{Limitations}\label{sec:limitations}

This section records the boundaries a reader should keep in mind when interpreting our results: where sampling is thin, what the environment deliberately simplifies, and how far the construct claim extends beyond this game. Every published run remains bit-replayable regardless; the items below bound interpretation, not reproducibility.

\begin{itemize}[leftmargin=2em]
\item \textbf{Three seeds bound the solo board, and the Arena is one
  world.} The solo results are means over three seeds, enough to show
  that adjacent models do not separate and that spread differs sharply
  by model (\S\ref{sec:campaign}), but not enough for inferential claims
  about any single pair. The Arena (\S\ref{sec:arena}) is one
  shared seed-7 world with no error bars, so its orderings within a few
  points should be read as ties, and aggregating several Arena worlds
  would call for a rank-based aggregate over per-world finishes, which
  we leave to future work.
\item \textbf{Scripted-opponent realism.} The 15 opponents are disciplined
  hand-written scripts by design (a fixed track is what makes absolute scores
  comparable), so the environment does not model adaptive rival managers; the
  Arena is the adversarial complement, not a realism claim.
\item \textbf{Construct validity.} FM-Bench instantiates long-horizon
  management in one domain. The four demands are domain-general design
  targets, but transfer of FM-Bench rankings to other long-horizon
  settings is untested.
\item \textbf{Behavioral metrics are correlational.} The
  capability decomposition (\S\ref{sec:capabilities}) reads reproducible
  metrics off the runs; a metric can be confounded by other traits
  (the warning-exposure metric demonstrably is), coefficients over $n = 15$
  are indicative, and the causal probes that would upgrade them
  (notebook ablation, plan injection) are future work.
\end{itemize}

\section{Reproducibility}\label{sec:repro}

Deterministic engine (counter-based RNG keyed by (domain, entity, day); fixed daily phase order; no wall clock); a 284-test suite including golden-hash behavior guards, full-tool-path replay bit-identity, crash-resume equivalence, AST truth-isolation, observation-audit, and calibration regressions; every result file carries engine commit, params hash, score version, and the full knob set; scripted baselines and the full calibration suite run for free on a laptop (20 simulated years in 76\,s single-core). Open Track submissions are HMAC-signed and re-verified by server-side replay. Scored worlds are salted (effective seed $=$ HMAC-SHA256 of a secret salt and a public nonce), keeping worlds unguessable despite the open engine; effective seeds are disclosed after each round closes, so every run remains fully replay-auditable. The reported runs exercised this machinery: several 20-year runs survived provider outages or credit exhaustion mid-run and resumed losslessly from the fsync'd action log (no completed turn was re-purchased), and every reported result is bit-replayable from its log, with full per-turn transcripts retained in the repository.

\section{The Roster}\label{app:roster}

The roster (Table~\ref{tab:roster}) fixes four selection principles even as individual model IDs rotate: within-family tier curves, closed against open frontier under identical rules, an agentic and tool-use specialist, and a disciplined scripted anchor in the 16th seat. All 15 serving IDs were validated by tool-calling pings before launch. Same-role substitution is allowed in later rounds, but the principles are fixed.

\begin{table}[t]
\centering
\small
\caption{The roster: 15 frontier LLM seats plus one scripted
anchor, July 2026.}
\label{tab:roster}
\begin{tabular}{@{}rlll@{}}
\toprule
\# & seat & provider & role \\
\midrule
1 & \texttt{claude-fable-5}~\citep{anthropic2026fable5} & Anthropic & new flagship \\
2 & \texttt{claude-opus-4.8}~\citep{anthropic2026opus48} & Anthropic & previous flagship \\
3 & \texttt{claude-haiku-4.5}~\citep{anthropic2025haiku45} & Anthropic & small tier \\
4 & \texttt{gpt-5.6-sol}~\citep{openai2026gpt56} & OpenAI & new flagship \\
5 & \texttt{gpt-5.6-terra}~\citep{openai2026gpt56} & OpenAI & balanced tier \\
6 & \texttt{claude-sonnet-5}~\citep{anthropic2026sonnet5} & Anthropic & workhorse mid (4-point family curve) \\
7 & \texttt{gemini-3.5-flash}~\citep{google2026gemini35flash} & Google & latest GA flagship \\
8 & \texttt{gemini-3-flash}~\citep{google2025gemini3flash} & Google & previous-gen fast \\
9 & \texttt{deepseek-v4-pro}~\citep{deepseek2026v4} & Together & open flagship \\
10 & \texttt{grok-4.5}~\citep{xai2026grok45} & xAI & frontier \\
11 & \texttt{kimi-k2.6}~\citep{moonshot2026kimik26} & Together & open agentic specialist \\
12 & \texttt{qwen3.7-max}~\citep{alibaba2026qwen37max} & Together & open flagship \\
13 & \texttt{glm-5.2}~\citep{zhipu2026glm52} & Together & open top tier \\
14 & \texttt{muse-spark-1.1}~\citep{meta2026musespark} & Meta Model API & Meta agentic model \\
15 & \texttt{minimax-m3}~\citep{minimax2026m3} & Together & open flagship \\
16 & heuristic & --- (scripted) & disciplined-script anchor \\
\bottomrule
\end{tabular}
\end{table}

\section{What the Human Players Reported}\label{app:humanfeedback}

After their runs, we asked the six players (\S\ref{sec:human}) what they found hard and what they would change. They were not shown the four demands, the capability decomposition, or any model result. We group their reports below and note, where one exists, the model-side measurement that matches.

\paragraph{Hidden information, reported as a low tolerance for error.} Players could not see what a seller would accept or what a player would re-sign for, so a misjudged offer was unrecoverable rather than correctable. Two of them settled on age heuristics, releasing or not renewing past 30, because the decline curve is unobservable. One reported angering sellers with low offers and paying more afterward, which is the repeat-pair markup of the counter-adaptive market (\S\ref{sec:loop}) felt from the inside. Lineup choice raised the same problem: no player was sure whether to field the highest ability, the best current form, or the freshest legs, and one switched from the first to the last over the run.

\paragraph{Cumulative consequences, reported as delayed bills.} Clicking through the early stops felt free, because the engine fields a short squad rather than refusing to continue, and one player took that as a signal that buying was optional; the losses arrived seasons later. This is the idle-cash pattern of \S\ref{sec:capabilities}, the strongest negative correlate among models, reached by a different route. A second delayed bill came from squad depth. One player released weak low-wage players to shorten a long list and found later that those were exactly the squad depth needed once the first eleven tired or got injured, and revised the policy for the rest of the run. Two further timing effects went unnoticed until late: the transfer market thins out as seasons pass, so buying early is worth more than it looks, and academy players are cheap but start low and mature slowly, which left one player unsure whether youth investment ever repaid. The models leave the same question open: the academy-harvest metric that looked informative on one seed does not survive three (Appendix~\ref{app:behavior}).

\paragraph{Credit assignment, reported as helplessness.} The most frequent complaint was not knowing which decision caused a losing season. One player fielded the strongest available eleven and lost repeatedly, then won with a weaker one after injuries forced changes, and could not tell whether formation, opponent matchup, or noise explained it. Once board confidence began to fall, players described having no idea what to change, and four of the six were fired.

\paragraph{Multi-objective pressure, reported as multitasking.} Players described the pressure as running valuation and squad maintenance at once, pricing targets in the market while watching for tired players to substitute, with a wage bill above 85\% of revenue leaving them one bad season from dismissal.

\paragraph{What the players did that the models do not.} Three behaviors stand out against the model-side profiles. One player derived a rule that always holds from two mechanisms: sign for the maximum five years because ability grows and money inflates, and apply it for the rest of the run; renewal lead time is the strongest positive correlate among models, but no model states the invariant. One revised a squad-depth policy in the middle of a run after seeing it fail, whereas in models mid-run tactic switching correlates negatively with score, so course changes there are usually thrash rather than repair. One wrote an external helper with an LLM to surface state the interface leaves implicit, an act of tool-building no model performed. The players also asked for a larger strategy space, in particular opponent-specific formations, after noticing that a weaker eleven in a better shape sometimes wins.

\paragraph{Reading these reports.} They are six self-reports on one seed and are not evidence about capability rankings. We use them for two narrower purposes: as construct-validity evidence, since players who were never told the design named all four demands as the hard parts, and as design input, since the operational load they describe, tracking fatigue, contract clocks, and market depth by eye, is exactly what an agent absorbs at no cost.

\section{The Scripted Anchors}\label{app:anchors}

Three blind scripts run the same track as the models (\S\ref{sec:campaign}). None of them reads hidden state; all three go through the same 26-tool interface and per-stop budgets as an agent, and all three are deterministic, so they cost nothing to re-run and fix the scale at the bottom of the board the way the oracle fixes it at the top (Appendix~\ref{app:oracle}).

\paragraph{Heuristic, a disciplined manager.} The heuristic is a hand-written policy that plays the game the way a careful club would. It keeps the wage bill under 60\% of revenue, renews the young core and sells players past 32, buys young value targets when a window is open and, when wage headroom allows, promotes from the academy at intake, rotates the lineup by fatigue, switches playing style against the opponent's table position, and invests in facilities at season settlement when cash is ample. It answers incoming offers on a keep-or-sell rule, accepting good prices for players below the squad's median ability and countering at a markup otherwise. It is the baseline that matters: everything a manager should do mechanically, done without judgment about which season it is.

\paragraph{Idle, the floor.} The idle policy submits nothing at the draft and takes no action at any stop thereafter. The engine keeps the club running with default handling, so idle measures what the world does to a club whose manager never intervenes.

\paragraph{Random, below the floor.} The random policy answers pending offers by coin flip, then takes zero to three arbitrary actions per stop, drawn from bidding a random amount on a random target, listing or unlisting a random player, and setting a random formation and style. Its draws are seeded from the world seed and the stop ID, so the run is reproducible. Random scores below idle, which is itself informative: in this environment uninformed activity destroys more value than inactivity.

\section{Opponent Tiers}\label{app:tiers}

The 15 scripted opponents of the solo track (\S\ref{sec:campaign}) are one manager implementation under three settings of five competence levers (Table~\ref{tab:tiers}). Medium is that implementation with every lever neutral, which is exactly the untiered scripted manager, and easy and hard turn the same five levers down or up. Every lever is a deterministic function of state, so tiers add no randomness, and none of them touches the observation path, so an easy board is worse at judging players rather than being differently informed.

\begin{table}[t]
\centering
\small
\caption{The five competence levers. Medium is the neutral setting.}
\label{tab:tiers}
\begin{tabularx}{\textwidth}{@{}Xrrr@{}}
\toprule
lever & easy & medium & hard \\
\midrule
scouting noise, multiplier on the belief sigma & 1.3 & 1.0 & 0.35 \\
lineup rotates for fatigue & no & yes & yes \\
market days acted on & every 2nd & every & every \\
gap required before buying an upgrade & 1.5 & 1.0 & 0.25 \\
wage budget, multiplier on the board target & 0.92 & 1.0 & 1.10 \\
\bottomrule
\end{tabularx}
\end{table}

Tiers are assigned by a frozen cyclic hard, medium, easy pattern over the club slots ranked by initial cash, so each wealth band contains all three tiers and money is decorrelated from competence. The pattern is a pure function of the parameters, with no random draw, so every tested model meets the same 15 opponents in the same slots.

An earlier calibration of the levers came out net easier than no tiers at all, because noisy easy sellers priced players badly enough to be free bargains and their unrotated first elevens donated points across the league, while the hard setting was too subtle to compensate. The current values temper how exploitable easy is and give hard sharper beliefs, proactive buying, and a more ambitious wage budget.

\section{Arena Data Quality: the Notebook Audit}\label{app:audit}

An LLM auditor read every model's self-authored notebook writes across the first seven game-years of the Arena run (\S\ref{sec:arena}). All 15 models stay self-consistent and factually grounded: 10 clean, 5 with minor style or hygiene issues, none broken. No model hallucinated its squad, invented finances, or lost track of its league position, and weak-but-coherent play, meaning verbose crisis vocabulary, boilerplate, doom loops, and an aging-squad decline, is distinguishable from broken play. Model play and the interface are frozen and identical to the 1v15 solo track: the same engine, the same 26-tool interface, and the same rules, so the two tracks stay comparable. The audit also surfaced two interface-friction phenomena, a standing list order that reads as consent to auto-sell and the money unit-scale duality, which are logged as candidates for a future revision to be applied to both tracks together or not at all.

\section{The Oracle Policy}\label{app:oracle}

This appendix details the privileged oracle summarized in \S\ref{sec:protocol}. Its privilege is information, not power. It reads true hidden state straight from the engine, the one thing no legitimate agent can do, but every action goes through the same tool interface, the same per-stop budgets, and the same legality checks as any agent. Because the engine's random streams are pure functions of the seed, entity, and day, even the draw that resolves a same-tick negotiation is readable truth, so each exploit below is a deterministic read of world state rather than a search.

\begin{itemize}[leftmargin=2em]
\item \textbf{Transfer arbitrage.} The oracle scans every player in the
  league and computes each seller's exact accept threshold, the asking
  price adjusted by the seller's style and distress, then bids exactly
  that. Every purchase closes on the first offer at the lowest price
  the seller would have taken. Buys are ranked by true worth against
  effective cost, including the drag that unamortized fee liabilities
  put on the score.
\item \textbf{Exact-threshold contracts.} Renewals and free-agent
  signings offer the player's exact reservation wage, so no negotiation
  round is wasted, no cooldown is triggered, and the wage bill is the
  lowest in the league. Since an accepted offer resets the wage,
  overpaid contracts are cut, and because contract offers sit outside
  the per-stop negotiation budget while every accepted renewal adds
  morale, the best players are re-renewed at their known reservation
  wage whenever that costs nothing, keeping the squad's morale multiplier
  near its cap.
\item \textbf{Academy sniping.} Only players with true high potential
  are promoted, at the last stop before their academy terms lapse,
  since academy development outpaces bench development. Academy and
  training facilities are funded early so that intake quality and
  growth compound.
\item \textbf{Peak-timed sales.} Every incoming bid is countered at the
  buyer's exact ceiling, the lower of its transfer budget and 1.15
  times its believed value, quantized to the price ladder. Surplus and
  post-peak players therefore sell for the most any buyer in the world
  would pay, before the deterministic decline curve bites. Resales
  within a year of purchase are refused so that the counter-adaptive
  market never taxes the flywheel.
\item \textbf{Zero-waste scheduling.} The starting eleven is re-solved
  at every stop from true ability, form, morale, and projected fatigue
  across all three formations, and the playing style counter-picks the
  known styles of the upcoming opponents, net of the style-switch
  penalty.
\end{itemize}

The policy is also score-aware. Trading profit and squad growth land in the net-worth channel, sales never hollow the last-three-season window, excess cash discounted by the score is recycled into facilities and squad assets, and contracts are renewed so that nobody expires off the books before the final snapshots. A survival controller reads board confidence directly. Below an emergency threshold, set well above the engine's own vote and firing thresholds, every margin discipline yields to immediate on-pitch strength, so the oracle completes the horizon on every seed. Engine code may never import the oracle module, which a CI isolation test enforces, and oracle scores are reported alongside the ladder but never mixed into it.

\section{Properties of the Composite Score}\label{app:score}

This appendix collects four elementary properties of the composite score
(\S\ref{sec:scoring}).
Write $x^{+} = \max(x, 0)$ and $x^{-} = \min(x, 0)$, so that
\begin{align*}
S_{\text{raw}}(H, \mathit{VA}, M) \;=\;{}& w_h \ln\!\bigl(1 + H^{+}/h_d\bigr)
  + w_v \operatorname{sign}(\mathit{VA})
    \ln\!\bigl(1 + |\mathit{VA}|/v_d\bigr)
  + w_m \ln\!\bigl(1 + M^{+}/m_d\bigr),\\
S_{\text{final}} \;=\;{}& \rho \cdot S_{\text{raw}}^{+} + S_{\text{raw}}^{-},
\qquad
\rho =
\begin{cases}
1 & \text{run completed},\\[2pt]
\rho_0 \, (t/T)^{\rho_1} & \text{settled early at } t \in [0, T),
\end{cases}
\end{align*}
with weights $(w_h, w_v, w_m) = (18, 10, 6)$, divisors
$(h_d, v_d, m_d) = (60, 40, 80)$, and $\rho_0 = 0.8$, $\rho_1 = 1.5$.

\medskip \noindent\textbf{Proposition A.1 (no fire-to-cap arbitrage).} \emph{For fixed $S_{\text{raw}}$, the map $t \mapsto S_{\text{final}}$ is nondecreasing on $[0, T]$, and $S_{\text{final}} \leq S_{\text{raw}}$ with equality iff the run completes or $S_{\text{raw}} \leq 0$.}

\smallskip \noindent\emph{Proof.} If $S_{\text{raw}} \leq 0$ then $S_{\text{raw}}^{+} = 0$ and $S_{\text{final}} = S_{\text{raw}}$ for every $t$: the discount multiplies only the positive part, and the negative part passes through undiscounted, so early settlement leaves a negative score exactly unchanged: ``get fired early to cap losses'' changes nothing. If $S_{\text{raw}} > 0$ and the run settles early at $t < T$, then $S_{\text{final}} = \rho_0 (t/T)^{\rho_1} S_{\text{raw}}$ with $\partial S_{\text{final}} / \partial t = \rho_0 \rho_1 t^{\rho_1 - 1} T^{-\rho_1} S_{\text{raw}} > 0$, and $\rho_0 (t/T)^{\rho_1} \leq \rho_0 < 1$, while completion gives $\rho = 1$. Hence $\partial S_{\text{final}} / \partial t \geq 0$ throughout, and triggering an earlier settlement can only lower the score: it is never optimal. \hfill$\square$

\smallskip \noindent\emph{Remark.} The proposition freezes $S_{\text{raw}}$; the dynamic variant, whether $\rho(t) \cdot S(t)$ along a real trajectory can exceed the expected completed score, is exactly the self-fire exercise point probed empirically in CI. The discount is deliberately discontinuous at the horizon: $\rho \to \rho_0 = 0.8$ as $t \to T^{-}$ while completion pays $\rho = 1$, so surviving the final stretch carries a 25\% premium over settling just short of it; the jump points in the direction an agent cannot exploit.

\medskip \noindent\textbf{Proposition A.2 (channel monotonicity).} \emph{$S_{\text{raw}}$ is strictly increasing in $H$ on $H \geq 0$ (weakly on all of $\mathbb{R}$: the floor makes the honors term constant for $H < 0$), strictly increasing in $\mathit{VA}$ on all of $\mathbb{R}$, and weakly increasing in $M$ (strictly on $M \geq 0$).}

\smallskip \noindent\emph{Proof.} For $H \geq 0$ the honors term is $w_h \ln(1 + H/h_d)$ with derivative $w_h / (h_d + H) > 0$. For the VA term let $g(x) = \operatorname{sign}(x) \ln(1 + |x|/v_d)$: for $x \neq 0$, $g'(x) = 1/(v_d + |x|) > 0$, and at $x = 0$ both one-sided difference quotients converge to $1/v_d$, so $g$ is differentiable on $\mathbb{R}$ with $g' > 0$ everywhere, hence strictly increasing. The $M$ term is constant for $M < 0$ and has derivative $w_m / (m_d + M) > 0$ for $M \geq 0$. \hfill$\square$

\medskip \noindent\textbf{Proposition A.3 (fair-price transfer neutrality on the VA channel).} \emph{Net worth is booked as $N = C + \lambda Q + F - L$, with $C$ cash, $Q$ squad value, $\lambda = 0.8$ the squad liquidity discount, $F$ facilities, and $L$ outstanding fee liabilities (\S\ref{sec:scoring}). Buying a player of market value $v$ at price $p \geq v$, under any split of up-front cash and installments, changes net worth by $\Delta N = -p + \lambda v \leq -(1 - \lambda)\, v < 0$. Hence purchases at (or above) fair price strictly decrease net worth, and by Proposition A.2 the VA term weakly decreases: squad-shopping cannot pump the VA channel.}

\smallskip \noindent\emph{Proof.} The paid portion of $p$ leaves $C$ and the unpaid portion enters $L$, decreasing $C + \cdots - L$ by exactly $p$; the player enters $Q$ at value $v$, booked at $\lambda v$. So $\Delta N = -p + \lambda v$, which at $p = v$ equals $-(1 - \lambda) v < 0$ and is smaller still for $p > v$. Only buying below the discount wedge ($p < \lambda v$) raises $N$, the intended reward for scouting genuine bargains. Wage commitments add future outflows and only strengthen the inequality. \hfill$\square$

\smallskip \noindent\emph{Two scope notes.} (i) Cash above $3\times$ the annual wage bill is booked at $0.3$ (anti-hoarding, \S\ref{sec:scoring}); spending \emph{such penalized} cash on players can raise $N$ ($\lambda = 0.8 > 0.3$). This is the intended incentive to deploy hoarded cash, not an arbitrage: a purchase never books more than $\lambda < 1$ of the cash it consumes, so no purchase sequence can push $N$ above what the same cash counted at par would have given. (ii) The $M$ channel is \emph{absolute} squad value, so a purchase does raise the $M$ term. Protection of the \emph{total} score against squad-shopping therefore rests on $M$'s low weight ($6$ vs $10$) and the discount wedge $(1 - \lambda)$; it is an empirical claim (a per-channel value-added variant that removed the wedge resurrected this exploit and was rejected), not a theorem.

\medskip \noindent\textbf{Proposition A.4 (diminishing returns; no runaway channel).} \emph{Each channel term is concave in its input on the gain domain, so the marginal value of the $k$-th title (or the marginal million) is strictly decreasing, and no single channel can run away.}

\smallskip \noindent\emph{Proof.} $\frac{d^2}{dx^2} \ln(1 + x/d) = -(d + x)^{-2} < 0$ on $x \geq 0$, which covers the honors and squad terms; the VA term is odd, hence concave on $\mathit{VA} \geq 0$ and (symmetrically) convex on $\mathit{VA} \leq 0$, so gains and losses are compressed at the same rate. Concretely, the $k$-th league title ($+100$ honors points) is worth $w_h \ln\!\bigl((h_d + 100k) / (h_d + 100(k-1))\bigr)$, i.e.\ $17.7$, $8.7$, $5.9, \ldots$ points for $k = 1, 2, 3, \ldots$ Because each term grows only logarithmically in its input, overcoming a fixed additive lead on one channel requires a \emph{multiplicative} resource gap on that channel: the score is uncapped (the benchmark cannot saturate) yet no channel can dominate the composite. \hfill$\square$

\paragraph{A loose upper bound.} The channels are uncapped, so the composite has no mathematical supremum, but an impossibility bound follows from channel values no run can reach. Winning the title in all 20 seasons gives $H = 2000$ and an honors term of $18\ln(1 + 2000/60) = 63.7$. Draining the entire world's cash into value added, $\mathit{VA} = 10{,}000$M, gives $10\ln(1 + 10000/40) = 55.3$, and a full squad of maximum-ability players, $M \approx 6{,}000$M, gives $6\ln(1 + 6000/80) = 25.6$. With $\rho = 1$ the terms sum to ${\approx}145$. The world holds neither that much extractable cash nor that many such players, so the bound is loose by construction, and the oracle's 95.54 and the best blind model's 90.94 sit at roughly two-thirds of it.

\section{Behavioral Analysis: Definitions, Construction, and Additional Figures}\label{app:behavior}

This appendix documents the measurement pipeline behind \S\ref{sec:capabilities}. Everything is reproducible from the self-contained analysis folder published with the repository (scripts, mined data, and figures, with the extraction commands documented).

\paragraph{Replay gate.} All behavioral metrics are mined by bit-identical replay of each run's seed and action log for the 15 solo runs and the Arena archive. Two replay passes were run (one for activity/notebook/tactics, one reading engine truth, namely contract months remaining, player age and potential, and club cash, at the instant each action executed). Each pass verifies that the replayed final score reproduces the published $S_{\text{final}}$ of the run before its log is trusted; all 30 replays matched with zero drift. Solo action timing is therefore replay-exact; Arena timing maps each seat's stop index linearly onto its tenure (${\sim}19$ stops per club-year) and is approximate.

\paragraph{Metric definitions.} One item per capability of
\S\ref{sec:capabilities}; each model contributes the mean of its
statistic over the three seeds, and the stated direction is the one the
matrix axis rewards.
\begin{itemize}[leftmargin=2em]
\item \emph{Endgame awareness}: long-horizon actions (\texttt{invest}
  $+$ \texttt{promote\_youth}) per season; the statistic is the
  late-minus-mid contrast, the rate over years 17--20 minus the rate
  over years 2--16 (low is better).
\item \emph{Credit assignment}: idle-cash ratio, season-end cash $\div$
  net worth, averaged over seasons 2--20 (low is better). Net worth
  books excess cash at a discount (\S\ref{sec:scoring}), so values above
  100\% indicate an idle pile exceeding the club's entire discounted
  worth.
\item \emph{Proactive control}: renewal lead, the contract months
  remaining when a renewal negotiation \emph{episode} opens; offers to
  the same player within 60 game-days collapse into one episode, own-club
  players only; the statistic is the median over episodes (high is
  better). \emph{Last-minute share}: episodes opened at $\leq 6$ months.
\item \emph{Price discovery}: offers per completed signing, successful
  \texttt{make\_transfer\_offer} submissions $\div$ senior-squad
  arrivals with a positive recorded fee (drafted and promoted players
  excluded; low is better). The oracle's first-offer policy fixes the
  reference at 1.0 (Appendix~\ref{app:oracle}).
\item \emph{Memory curation}: each model's notebook is reconstructed at
  every season end under exact engine semantics (append joins with a
  newline, rewrite replaces); consecutive season-end states are compared
  by TF--IDF (1--2\, gram) cosine similarity, and the per-model mean is
  the consistency statistic. The matrix axis scores the distance
  $|s - 0.35|$ from the empirically best band (low is better). TF--IDF
  is chosen over neural embeddings for determinism and zero external
  dependencies; a semantic-embedding replication is future work.
\item \emph{Compute}: tokens by manifest accounting, input $+$ output
  $+$ cache read $+$ cache write from each run's manifest. The
  correlation in \S\ref{sec:capabilities} uses the total; the matrix
  axis scores efficiency, score per million tokens (high is better).
\end{itemize}

\begin{figure}[t]
\centering
\includegraphics[width=\textwidth]{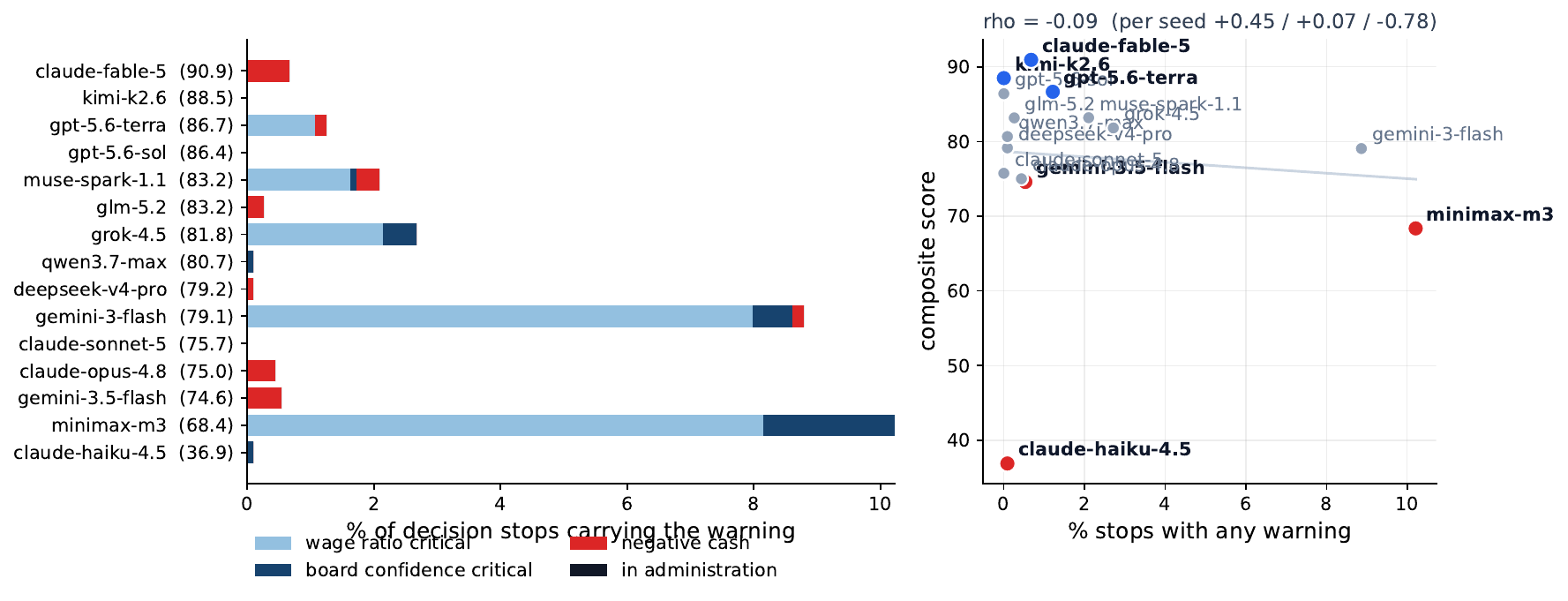}
\caption{The failed metric: share of stops carrying an engine warning,
by type (left) and against final score (right), means over the three
seeds. The per-seed coefficient runs $+0.45$ / $+0.07$ / $-0.78$ and
averages to nothing, so warning exposure is not reported as a
capability. Zero warnings conflates genuine anticipation with
do-nothing conservatism.}
\label{fig:warnings}
\end{figure}

\begin{figure}[t]
\centering
\includegraphics[width=\textwidth]{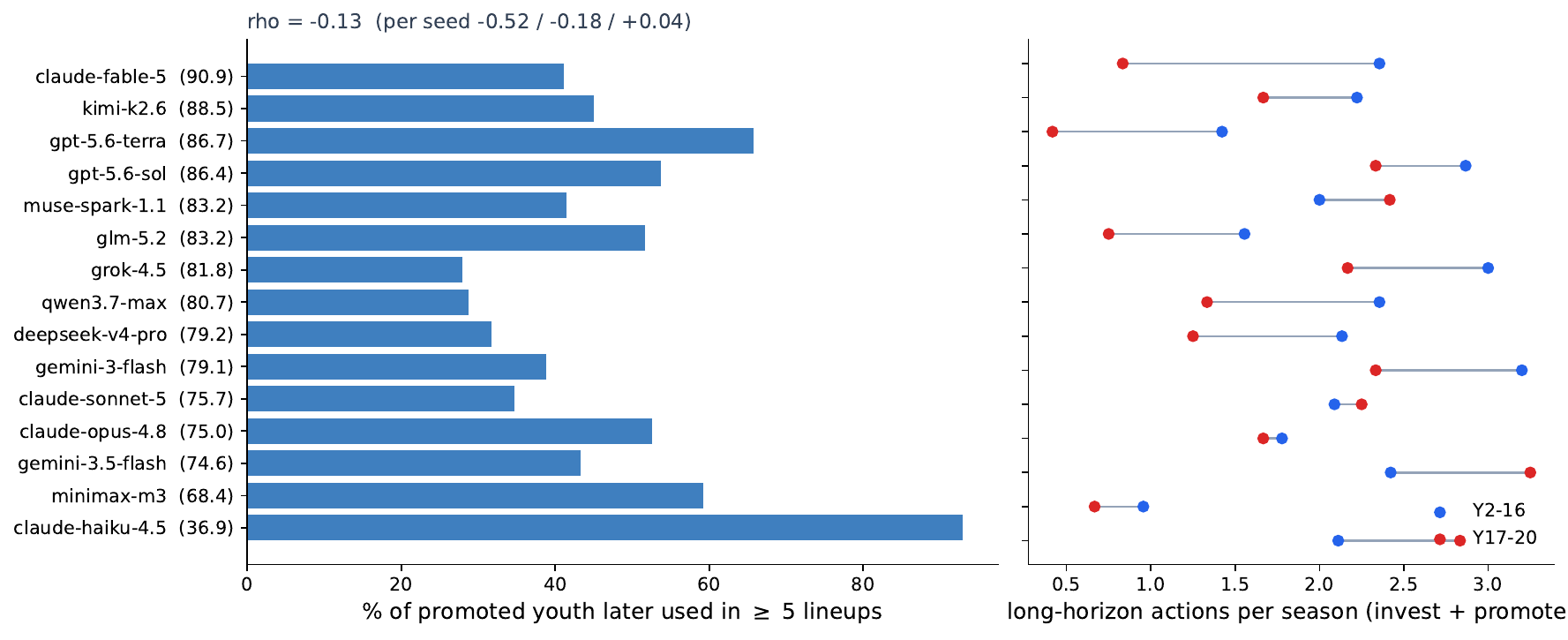}
\caption{Both panels are means over the three seeds. Left: youth
harvest rate (share of promotions later fielded in $\geq 5$ lineups);
the correlation that looked informative on seed 1 does not survive
three ($-0.52$ / $-0.18$ / $+0.04$). Right: endgame shift in
long-horizon actions per season, years 2--16 (blue) vs.\ 17--20 (red);
eleven models reduce late investment and four raise it.}
\label{fig:harvest}
\end{figure}

\paragraph{Failed metrics, kept in the record.}
\begin{itemize}[leftmargin=2em]
\item \emph{Warning exposure}: fraction of decision stops whose digest
  carried at least one engine warning (wage ratio, negative cash, board
  confidence, administration); per-seed sign unstable
  (Figure~\ref{fig:warnings}).
\item \emph{Youth harvest}: share of \texttt{promote\_youth} players later
  fielded in $\geq 5$ lineups; per-seed sign unstable
  (Figure~\ref{fig:harvest}).
\item \emph{Tactic reversals}: \texttt{set\_tactics} calls whose formation
  or style differs from the currently active pair (re-issuing the current
  tactic does not count); dropped as a capability axis for the same
  reason, though the raw spread stays wide, zero reversals for the winner
  in all three worlds against 31 on average for
  \texttt{claude-haiku-4.5}.
\end{itemize}

\paragraph{Token-file erratum.} A summary token file that circulated with early drafts mixed in token totals from earlier runs for ten of the fifteen models. All token numbers in this paper are recomputed from the run manifests; the superseded file is retained in the repository for audit.

\paragraph{Capability-matrix construction (Figure~\ref{fig:capmatrix}).} Each column rank-normalizes one metric across the 15 models to $[0, 1]$ (1 = best rank; ties broken by value order), in the direction stated in the definitions above. Rank scores are a visualization-grade aggregation over $n = 15$ models; the evidentiary weight rests on the underlying metrics.

\begin{table}[t]
\centering
\small
\caption{Per-model profiles over the three seeds. Score is the mean and SD
the standard deviation of $S_{\text{final}}$; axis extremes come from the
matrix of Figure~\ref{fig:capmatrix}.}
\label{tab:profiles}
\begin{tabularx}{\textwidth}{@{}lrr>{\raggedright\arraybackslash}X@{}}
\toprule
seat & score & sd & profile \\
\midrule
\texttt{claude-fable-5} & 90.9 & 5.2 & generalist, no axis below 0.79; best bidder in the field at 9 offers per signing and the sharpest endgame reduction \\
\texttt{kimi-k2.6} & 88.5 & 0.1 & the steadiest model in the campaign, scoring within a third of a point on three different worlds \\
\texttt{gpt-5.6-terra} & 86.7 & 1.2 & minimalist: cheapest compute in the field (28M) and an early endgame reduction, but the shortest renewal leads among the top models \\
\texttt{gpt-5.6-sol} & 86.4 & 2.5 & archivist memory (similarity 0.91, append-only) and the weakest bidding of the top group (45 offers per signing) \\
\texttt{muse-spark-1.1} & 83.2 & 12.0 & strong on two worlds and 20 points weaker on the third; one of four models that raise long-horizon spending at the end \\
\texttt{glm-5.2} & 83.2 & 0.9 & stable across seeds but loose with cash (135\% idle ratio) \\
\texttt{grok-4.5} & 81.8 & 9.9 & compute as substitute: long renewal leads and heavy spend (167M tokens) with wide seed-to-seed swings \\
\texttt{qwen3.7-max} & 80.7 & 10.4 & longest renewal leads after the winner, undone by a 103\% idle ratio and churning memory (similarity 0.23) \\
\texttt{deepseek-v4-pro} & 79.2 & 2.7 & near-static notebook (0.81) and the heaviest spend in the field (194M tokens) \\
\texttt{gemini-3-flash} & 79.1 & 13.5 & best cash discipline in the field (46\%) but the least willing to cut endgame spending; collapses on the hardest seed \\
\texttt{claude-sonnet-5} & 75.7 & 5.0 & churning memory (0.20, the lowest) and no endgame reduction \\
\texttt{claude-opus-4.8} & 75.0 & 2.5 & shortest renewal leads in the field (10 months) with a 134\% idle ratio \\
\texttt{gemini-3.5-flash} & 74.6 & 11.0 & worst price discovery (73 offers per signing) and the largest endgame ramp-up \\
\texttt{minimax-m3} & 68.4 & 12.3 & prefers veteran signings, short leads, and swings 23 points across seeds \\
\texttt{claude-haiku-4.5} & 36.9 & 22.7 & weak on every axis at once: 196\% idle cash, 11-month leads, a rising endgame rate, and the widest spread in the campaign \\
\bottomrule
\end{tabularx}
\end{table}

\begin{figure}[t]
\centering
\includegraphics[width=\textwidth]{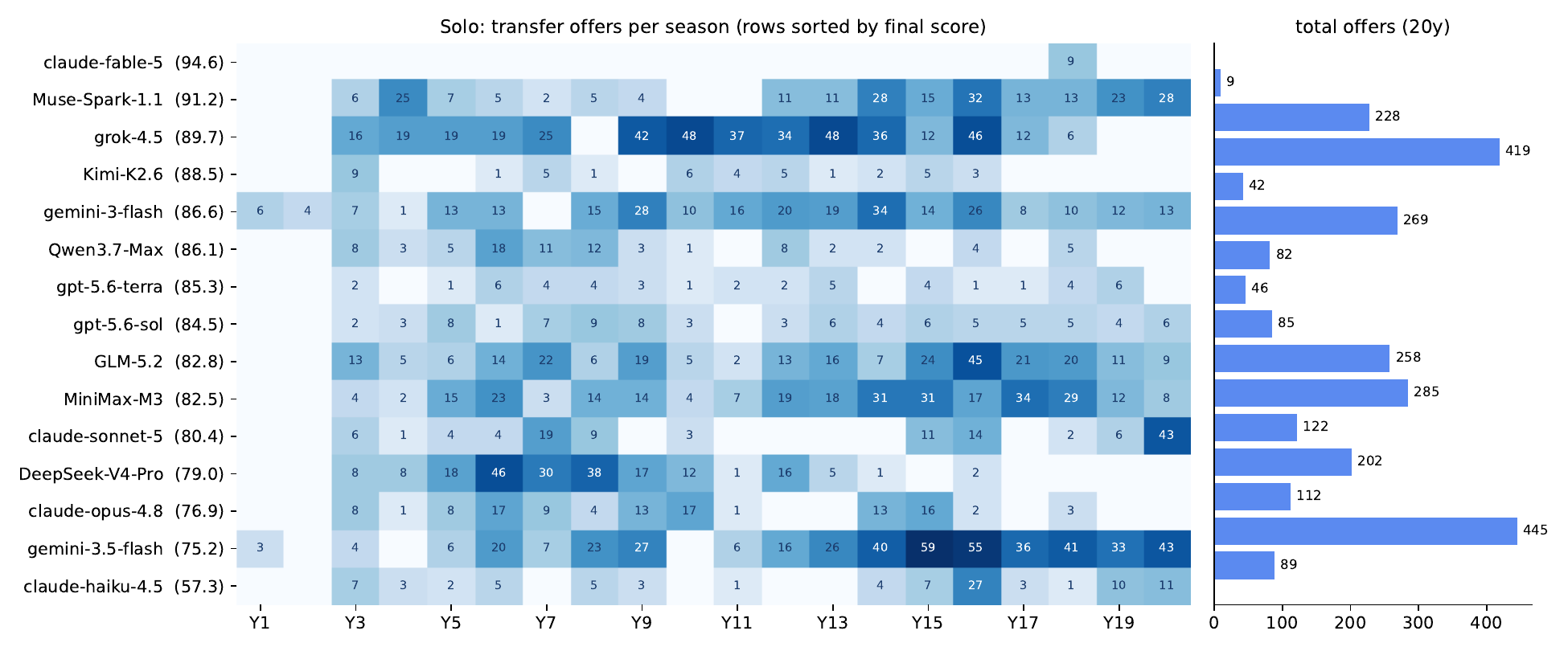}
\caption{Transfer offers per season, solo track (rows sorted by final
score; totals at right). The winner's 9 offers are all in the first two
seasons; the field-wide offer-to-completion conversion is 2--5\%.}
\label{fig:offers-solo}
\end{figure}

\begin{figure}[t]
\centering
\includegraphics[width=\textwidth]{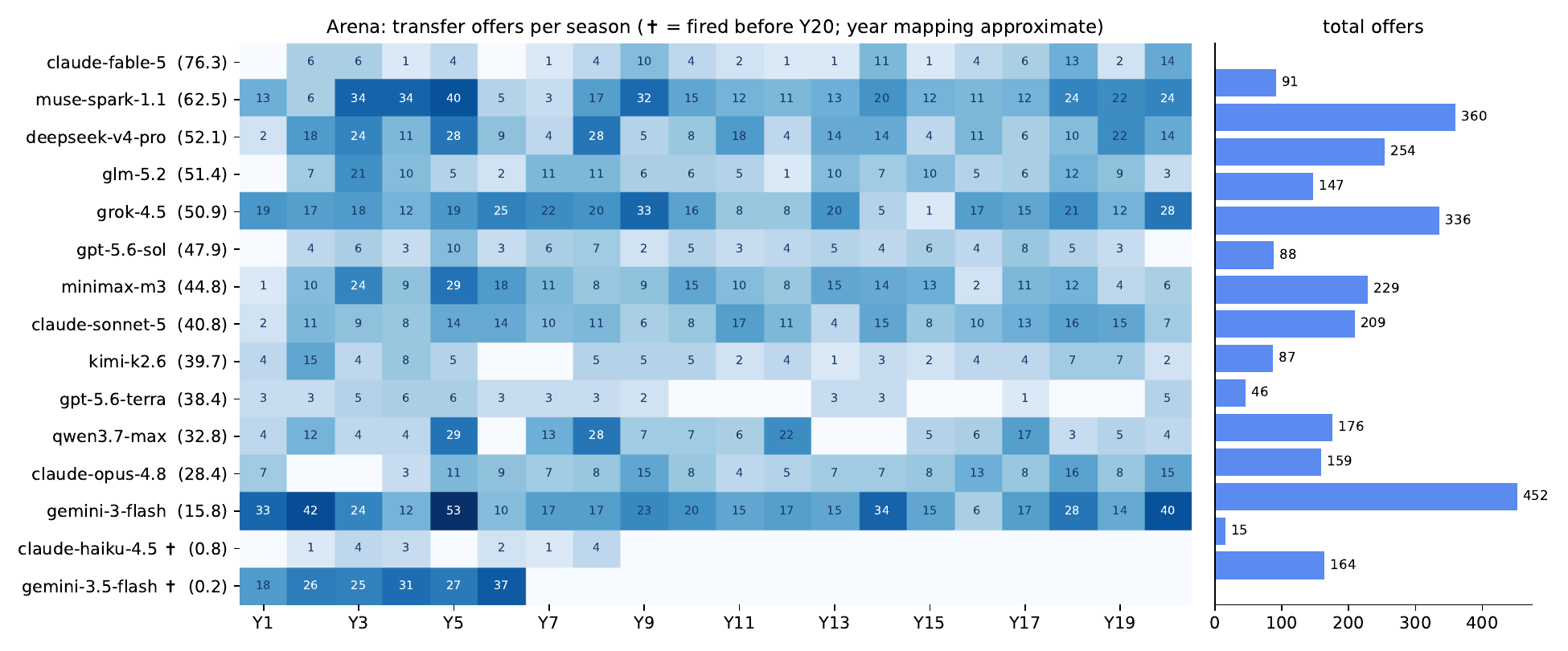}
\caption{Transfer offers per season in the Arena (approximate year
mapping; daggered models settled early). The winner's offer count rises from
9 (solo) to 91: the same model under a different, correct judgment of
market liquidity.}
\label{fig:offers-arena}
\end{figure}

\begin{figure}[t]
\centering
\includegraphics[width=0.7\textwidth]{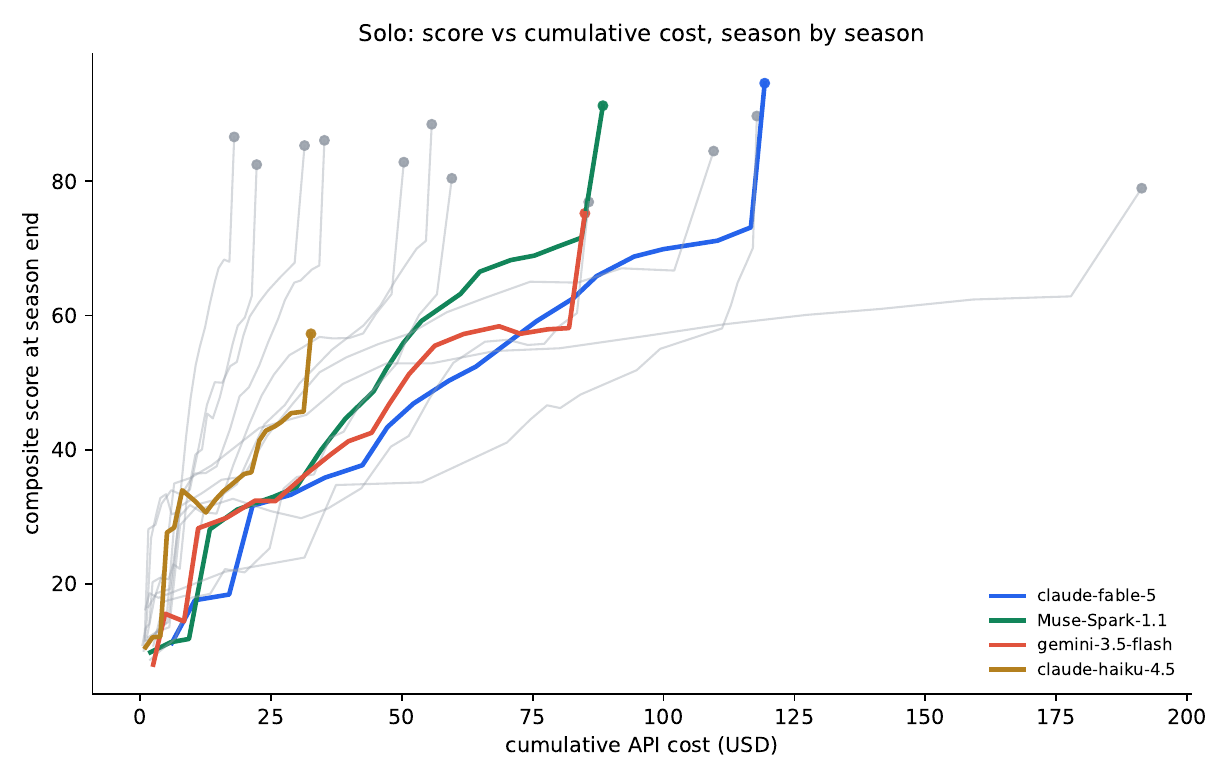}
\caption{Season-by-season cumulative API cost vs.\ composite score, solo
track. Steep curves convert dollars into score throughout; flat long curves
do not. The full 20-year runs span \$18 to \$191.}
\label{fig:costpaths}
\end{figure}


\begin{figure}[t]
\centering
\includegraphics[width=0.72\textwidth]{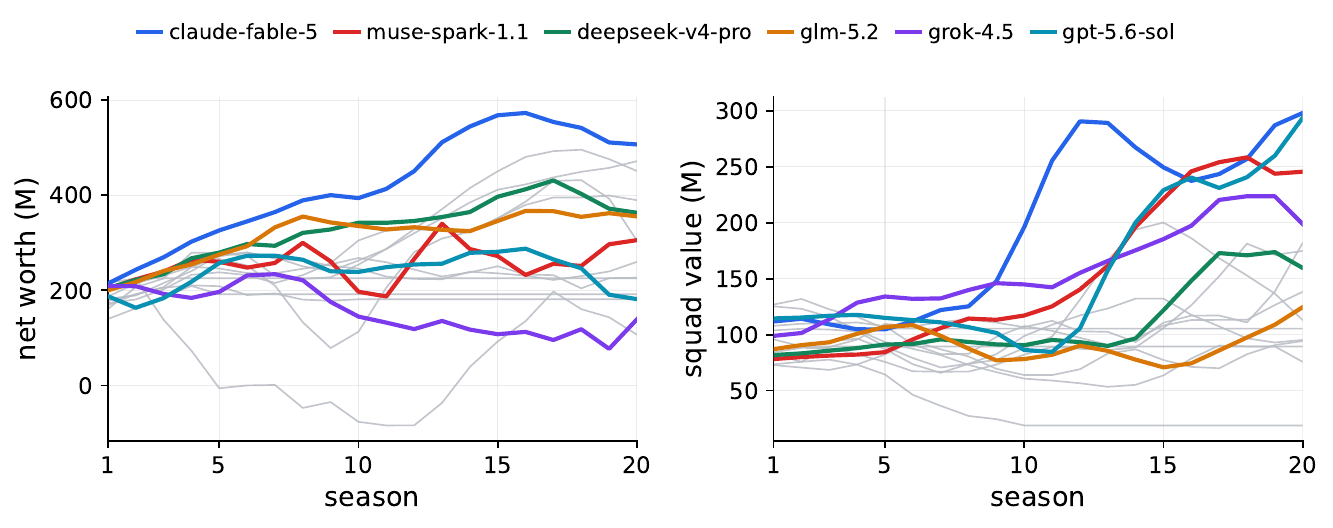}
\caption{The asset channels behind Figure~\ref{fig:arena-traj}: per-season
net worth and squad value for all 16 Arena seats, the six
highest-scoring seats colored as in Figure~\ref{fig:arena-traj} and the
grey lines the remaining ten (numeric snapshots in
Table~\ref{tab:snapshots-arena}). The winner's mid-table seasons are asset
accumulation (the net-worth curve keeps climbing while the standings do
not), and the conservative cash-holders' net worth grows on cash the
composite discounts.}
\label{fig:arena-assets}
\end{figure}

\paragraph{Season snapshots.} Tables~\ref{tab:snapshots-solo} and~\ref{tab:snapshots-arena} reproduce the numeric season snapshots of the public leaderboard page (years 5, 10, 15, 20). Cell format, matching the page: league position $\cdot$ cumulative honors points $\cdot$ net worth (M) $\cdot$ squad value (M) $\cdot$ completed-now composite score. They are the discrete complement of the trajectory figures (Figures~\ref{fig:solo-traj}, \ref{fig:arena-traj}, and~\ref{fig:arena-assets}).

\begin{table}[t]
\centering
\caption{Solo season snapshots (seed 1). Cell format:
position $\cdot$ honors $\cdot$ net worth $\cdot$ squad value $\cdot$
score. Rows sorted by final score.}
\label{tab:snapshots-solo}
\resizebox{\textwidth}{!}{%
\begin{tabular}{@{}lllll@{}}
\toprule
seat & Y5 & Y10 & Y15 & Y20 \\
\midrule
claude-fable-5 & 4th $\cdot$ 140H $\cdot$ 293M $\cdot$ 130M $\cdot$ 33.3 & 1st $\cdot$ 480H $\cdot$ 470M $\cdot$ 218M $\cdot$ 50.3 & 1st $\cdot$ 900H $\cdot$ 1048M $\cdot$ 443M $\cdot$ 65.9 & 1st $\cdot$ 1400H $\cdot$ 1742M $\cdot$ 634M $\cdot$ 94.6 \\
muse-spark-1.1 & 3rd $\cdot$ 120H $\cdot$ 302M $\cdot$ 158M $\cdot$ 31.1 & 1st $\cdot$ 460H $\cdot$ 468M $\cdot$ 196M $\cdot$ 48.7 & 1st $\cdot$ 960H $\cdot$ 1037M $\cdot$ 457M $\cdot$ 66.5 & 1st $\cdot$ 1460H $\cdot$ 1298M $\cdot$ 315M $\cdot$ 91.2 \\
grok-4.5 & 4th $\cdot$ 60H $\cdot$ 289M $\cdot$ 169M $\cdot$ 23.9 & 1st $\cdot$ 460H $\cdot$ 318M $\cdot$ 107M $\cdot$ 46.6 & 1st $\cdot$ 880H $\cdot$ 618M $\cdot$ 225M $\cdot$ 58.1 & 1st $\cdot$ 1380H $\cdot$ 1106M $\cdot$ 634M $\cdot$ 89.7 \\
kimi-k2.6 & 3rd $\cdot$ 120H $\cdot$ 344M $\cdot$ 116M $\cdot$ 31.9 & 1st $\cdot$ 440H $\cdot$ 430M $\cdot$ 99M $\cdot$ 46.7 & 1st $\cdot$ 940H $\cdot$ 668M $\cdot$ 276M $\cdot$ 61.4 & 1st $\cdot$ 1440H $\cdot$ 733M $\cdot$ 400M $\cdot$ 88.5 \\
gemini-3-flash & 3rd $\cdot$ 160H $\cdot$ 247M $\cdot$ 165M $\cdot$ 32.8 & 1st $\cdot$ 500H $\cdot$ 339M $\cdot$ 111M $\cdot$ 42.1 & 1st $\cdot$ 1000H $\cdot$ 675M $\cdot$ 400M $\cdot$ 61.5 & 1st $\cdot$ 1420H $\cdot$ 812M $\cdot$ 618M $\cdot$ 86.6 \\
qwen3.7-max & 3rd $\cdot$ 120H $\cdot$ 404M $\cdot$ 127M $\cdot$ 34.0 & 1st $\cdot$ 360H $\cdot$ 582M $\cdot$ 140M $\cdot$ 48.0 & 1st $\cdot$ 780H $\cdot$ 917M $\cdot$ 256M $\cdot$ 62.3 & 1st $\cdot$ 1200H $\cdot$ 1305M $\cdot$ 248M $\cdot$ 86.1 \\
gpt-5.6-terra & 3rd $\cdot$ 140H $\cdot$ 359M $\cdot$ 120M $\cdot$ 34.0 & 1st $\cdot$ 440H $\cdot$ 565M $\cdot$ 102M $\cdot$ 50.1 & 1st $\cdot$ 780H $\cdot$ 808M $\cdot$ 247M $\cdot$ 59.8 & 1st $\cdot$ 1280H $\cdot$ 987M $\cdot$ 186M $\cdot$ 85.3 \\
gpt-5.6-sol & 4th $\cdot$ 140H $\cdot$ 343M $\cdot$ 135M $\cdot$ 33.6 & 1st $\cdot$ 460H $\cdot$ 615M $\cdot$ 176M $\cdot$ 51.5 & 1st $\cdot$ 960H $\cdot$ 762M $\cdot$ 217M $\cdot$ 62.6 & 2nd $\cdot$ 1220H $\cdot$ 1109M $\cdot$ 274M $\cdot$ 84.5 \\
glm-5.2 & 3rd $\cdot$ 120H $\cdot$ 294M $\cdot$ 121M $\cdot$ 31.8 & 1st $\cdot$ 420H $\cdot$ 513M $\cdot$ 110M $\cdot$ 48.1 & 3rd $\cdot$ 760H $\cdot$ 712M $\cdot$ 58M $\cdot$ 56.6 & 1st $\cdot$ 1080H $\cdot$ 1331M $\cdot$ 266M $\cdot$ 82.8 \\
minimax-m3 & 10th $\cdot$ 20H $\cdot$ 324M $\cdot$ 154M $\cdot$ 20.9 & 1st $\cdot$ 240H $\cdot$ 492M $\cdot$ 100M $\cdot$ 39.3 & 1st $\cdot$ 540H $\cdot$ 827M $\cdot$ 155M $\cdot$ 51.8 & 1st $\cdot$ 1040H $\cdot$ 1245M $\cdot$ 327M $\cdot$ 82.5 \\
claude-sonnet-5 & 5th $\cdot$ 20H $\cdot$ 247M $\cdot$ 91M $\cdot$ 18.6 & 2nd $\cdot$ 180H $\cdot$ 363M $\cdot$ 97M $\cdot$ 36.0 & 1st $\cdot$ 480H $\cdot$ 647M $\cdot$ 94M $\cdot$ 50.2 & 1st $\cdot$ 980H $\cdot$ 1020M $\cdot$ 146M $\cdot$ 80.4 \\
deepseek-v4-pro & 5th $\cdot$ 140H $\cdot$ 385M $\cdot$ 124M $\cdot$ 35.8 & 1st $\cdot$ 480H $\cdot$ 638M $\cdot$ 111M $\cdot$ 52.9 & 1st $\cdot$ 820H $\cdot$ 732M $\cdot$ 110M $\cdot$ 58.7 & 3rd $\cdot$ 1080H $\cdot$ 996M $\cdot$ 163M $\cdot$ 79.0 \\
claude-opus-4.8 & 3rd $\cdot$ 140H $\cdot$ 308M $\cdot$ 100M $\cdot$ 32.7 & 1st $\cdot$ 280H $\cdot$ 467M $\cdot$ 78M $\cdot$ 40.5 & 2nd $\cdot$ 600H $\cdot$ 788M $\cdot$ 160M $\cdot$ 56.4 & 3rd $\cdot$ 760H $\cdot$ 1313M $\cdot$ 216M $\cdot$ 76.9 \\
gemini-3.5-flash & 2nd $\cdot$ 140H $\cdot$ 260M $\cdot$ 186M $\cdot$ 29.9 & 1st $\cdot$ 480H $\cdot$ 226M $\cdot$ 140M $\cdot$ 41.3 & 1st $\cdot$ 980H $\cdot$ 479M $\cdot$ 156M $\cdot$ 57.2 & 5th $\cdot$ 1040H $\cdot$ 877M $\cdot$ 170M $\cdot$ 75.2 \\
claude-haiku-4.5 & 12th $\cdot$ 100H $\cdot$ 293M $\cdot$ 132M $\cdot$ 28.4 & 7th $\cdot$ 200H $\cdot$ 395M $\cdot$ 39M $\cdot$ 33.8 & 3rd $\cdot$ 340H $\cdot$ 629M $\cdot$ 42M $\cdot$ 42.8 & 12th $\cdot$ 360H $\cdot$ 821M $\cdot$ 59M $\cdot$ 57.3 \\
\bottomrule
\end{tabular}}
\end{table}

\begin{table}[t]
\centering
\caption{Arena season snapshots (seed 7, one shared world);
the 15 models, as on the public page (the scripted anchor settled at
$t \approx 2.6$y). Cell format as in Table~\ref{tab:snapshots-solo}.
Seats that exhausted the revival cap keep their frozen score afterward,
visible as repeated values (\texttt{claude-haiku-4.5} from Y15,
\texttt{gemini-3.5-flash} from Y10). Rows sorted by final score.}
\label{tab:snapshots-arena}
\resizebox{\textwidth}{!}{%
\begin{tabular}{@{}lllll@{}}
\toprule
seat & Y5 & Y10 & Y15 & Y20 \\
\midrule
claude-fable-5 & 2nd $\cdot$ 180H $\cdot$ 326M $\cdot$ 105M $\cdot$ 47.4 & 5th $\cdot$ 240H $\cdot$ 394M $\cdot$ 196M $\cdot$ 56.4 & 5th $\cdot$ 440H $\cdot$ 568M $\cdot$ 250M $\cdot$ 71.3 & 3rd $\cdot$ 620H $\cdot$ 507M $\cdot$ 298M $\cdot$ 76.3 \\
muse-spark-1.1 & 8th $\cdot$ 0H $\cdot$ 260M $\cdot$ 84M $\cdot$ 18.5 & 7th $\cdot$ 40H $\cdot$ 197M $\cdot$ 117M $\cdot$ 24.0 & 2nd $\cdot$ 160H $\cdot$ 272M $\cdot$ 222M $\cdot$ 46.2 & 1st $\cdot$ 420H $\cdot$ 306M $\cdot$ 246M $\cdot$ 62.5 \\
deepseek-v4-pro & 9th $\cdot$ 0H $\cdot$ 279M $\cdot$ 91M $\cdot$ 20.0 & 10th $\cdot$ 140H $\cdot$ 342M $\cdot$ 91M $\cdot$ 44.5 & 4th $\cdot$ 200H $\cdot$ 397M $\cdot$ 122M $\cdot$ 52.2 & 5th $\cdot$ 200H $\cdot$ 363M $\cdot$ 160M $\cdot$ 52.1 \\
glm-5.2 & 4th $\cdot$ 40H $\cdot$ 275M $\cdot$ 107M $\cdot$ 30.3 & 1st $\cdot$ 160H $\cdot$ 335M $\cdot$ 78M $\cdot$ 46.1 & 8th $\cdot$ 200H $\cdot$ 346M $\cdot$ 71M $\cdot$ 49.3 & 6th $\cdot$ 200H $\cdot$ 356M $\cdot$ 125M $\cdot$ 51.4 \\
grok-4.5 & 7th $\cdot$ 20H $\cdot$ 197M $\cdot$ 134M $\cdot$ 21.1 & 6th $\cdot$ 260H $\cdot$ 145M $\cdot$ 145M $\cdot$ 40.0 & 7th $\cdot$ 320H $\cdot$ 108M $\cdot$ 185M $\cdot$ 36.4 & 8th $\cdot$ 520H $\cdot$ 139M $\cdot$ 199M $\cdot$ 50.9 \\
gpt-5.6-sol & 6th $\cdot$ 100H $\cdot$ 258M $\cdot$ 115M $\cdot$ 36.9 & 4th $\cdot$ 120H $\cdot$ 239M $\cdot$ 86M $\cdot$ 36.8 & 1st $\cdot$ 240H $\cdot$ 281M $\cdot$ 229M $\cdot$ 52.3 & 2nd $\cdot$ 280H $\cdot$ 182M $\cdot$ 294M $\cdot$ 47.9 \\
minimax-m3 & 3rd $\cdot$ 120H $\cdot$ 272M $\cdot$ 93M $\cdot$ 38.2 & 11th $\cdot$ 220H $\cdot$ 246M $\cdot$ 61M $\cdot$ 43.2 & 13th $\cdot$ 220H $\cdot$ 236M $\cdot$ 64M $\cdot$ 42.5 & 13th $\cdot$ 220H $\cdot$ 260M $\cdot$ 76M $\cdot$ 44.8 \\
claude-sonnet-5 & 11th $\cdot$ 100H $\cdot$ 246M $\cdot$ 82M $\cdot$ 35.9 & 13th $\cdot$ 100H $\cdot$ 268M $\cdot$ 64M $\cdot$ 36.4 & 9th $\cdot$ 100H $\cdot$ 251M $\cdot$ 77M $\cdot$ 36.0 & 10th $\cdot$ 160H $\cdot$ 226M $\cdot$ 94M $\cdot$ 40.8 \\
kimi-k2.6 & 13th $\cdot$ 0H $\cdot$ 238M $\cdot$ 76M $\cdot$ 18.1 & 8th $\cdot$ 20H $\cdot$ 305M $\cdot$ 82M $\cdot$ 26.9 & 3rd $\cdot$ 60H $\cdot$ 351M $\cdot$ 106M $\cdot$ 36.9 & 7th $\cdot$ 80H $\cdot$ 304M $\cdot$ 175M $\cdot$ 39.7 \\
gpt-5.6-terra & 5th $\cdot$ 40H $\cdot$ 266M $\cdot$ 108M $\cdot$ 28.8 & 2nd $\cdot$ 60H $\cdot$ 113M $\cdot$ 108M $\cdot$ 13.0 & 6th $\cdot$ 60H $\cdot$ 352M $\cdot$ 108M $\cdot$ 36.2 & 9th $\cdot$ 60H $\cdot$ 390M $\cdot$ 138M $\cdot$ 38.4 \\
qwen3.7-max & 14th $\cdot$ 20H $\cdot$ 279M $\cdot$ 87M $\cdot$ 19.2 & 12th $\cdot$ 20H $\cdot$ 254M $\cdot$ 88M $\cdot$ 13.6 & 12th $\cdot$ 180H $\cdot$ 451M $\cdot$ 200M $\cdot$ 34.2 & 11th $\cdot$ 180H $\cdot$ 451M $\cdot$ 113M $\cdot$ 32.8 \\
claude-opus-4.8 & 10th $\cdot$ 20H $\cdot$ 274M $\cdot$ 90M $\cdot$ 18.4 & 3rd $\cdot$ 40H $\cdot$ 263M $\cdot$ 99M $\cdot$ 21.2 & 10th $\cdot$ 40H $\cdot$ 412M $\cdot$ 132M $\cdot$ 27.8 & 14th $\cdot$ 40H $\cdot$ 472M $\cdot$ 95M $\cdot$ 28.4 \\
gemini-3-flash & 1st $\cdot$ 140H $\cdot$ $-6$M $\cdot$ 110M $\cdot$ 12.7 & 9th $\cdot$ 160H $\cdot$ $-77$M $\cdot$ 107M $\cdot$ 10.7 & 11th $\cdot$ 160H $\cdot$ 92M $\cdot$ 111M $\cdot$ 11.7 & 4th $\cdot$ 200H $\cdot$ 107M $\cdot$ 182M $\cdot$ 15.8 \\
claude-haiku-4.5 & 16th $\cdot$ 0H $\cdot$ 209M $\cdot$ 64M $\cdot$ 11.0 & 14th $\cdot$ 0H $\cdot$ 181M $\cdot$ 19M $\cdot$ 4.3 & 14th $\cdot$ 0H $\cdot$ 181M $\cdot$ 19M $\cdot$ 0.8 & 16th $\cdot$ 0H $\cdot$ 181M $\cdot$ 19M $\cdot$ 0.8 \\
gemini-3.5-flash & 12th $\cdot$ 0H $\cdot$ 192M $\cdot$ 105M $\cdot$ 2.1 & 15th $\cdot$ 0H $\cdot$ 192M $\cdot$ 105M $\cdot$ 0.2 & 15th $\cdot$ 0H $\cdot$ 192M $\cdot$ 105M $\cdot$ 0.2 & 12th $\cdot$ 0H $\cdot$ 192M $\cdot$ 105M $\cdot$ 0.2 \\
\bottomrule
\end{tabular}}
\end{table}

\section{The Tool Interface and the Decision-Stop Calendar}\label{app:tools}

This appendix lists the complete interface between agent and world: the 26 schema-generated tools (Table~\ref{tab:tools}), the decision-stop calendar that determines when the agent is asked to use them (Table~\ref{tab:stops}), the four memory layers (Table~\ref{tab:memory}), and the scale of one 20-year run (Table~\ref{tab:scale}). Query tools spend the per-stop query budget, negotiation actions spend the negotiation budget, and the notebook tools are free (\S\ref{sec:loop}); every UI button in the human web-play mode maps 1:1 to one of these tools.

\begin{table}[t]
\centering
\small
\caption{The 26-tool interface: 12 query tools, 11 action tools, 2
notebook tools, and 1 control tool. Descriptions follow the tool
schemas served to the agent.}
\label{tab:tools}
\begin{tabularx}{\textwidth}{@{}lX@{}}
\toprule
tool & what it does \\
\midrule
\multicolumn{2}{@{}l}{\emph{Query (12), charged to the 30-query
budget:}} \\
\texttt{get\_club\_overview} & the agent's club at a glance: division,
cash, board, facilities, tactics \\
\texttt{get\_squad} & the senior squad with ability bands, form,
contracts, wages \\
\texttt{get\_player} & detailed view of any player by id \\
\texttt{get\_league\_table} & league standings \\
\texttt{get\_fixtures} & the club's season fixtures and results \\
\texttt{get\_match\_report} & match report by match id \\
\texttt{get\_transfer\_market} & players available to buy: listed, free
agents, expiring contracts \\
\texttt{get\_finances} & cash, revenue, wage bill, active warnings \\
\texttt{get\_youth\_academy} & academy prospects with potential stars \\
\texttt{get\_history} & the run's archive: honors, seasons, transfers \\
\texttt{get\_inbox} & pending offers and open items \\
\texttt{get\_draft\_pool} & draft phase only: the shared template pool
with scouted ability bands, potential stars, prices, and preset
contracts, plus budget status \\
\midrule
\multicolumn{2}{@{}l}{\emph{Action (11); negotiation moves charged to
the 10-move budget:}} \\
\texttt{set\_lineup} & set the preferred starting XI (auto-completed if
players are unavailable) \\
\texttt{set\_tactics} & set formation and playing style from fixed
enums \\
\texttt{make\_transfer\_offer} & bid for another club's player; the
negotiation resolves within the stop \\
\texttt{respond\_to\_offer} & accept, reject, counter, or accept a
counter on a pending offer \\
\texttt{offer\_contract} & renew an own player or sign a free agent at
a wage and term \\
\texttt{list\_player} & put a player on or off the transfer list \\
\texttt{release\_player} & terminate a contract (severance: half of one
year's wage) \\
\texttt{promote\_youth} & promote an academy prospect to the senior
squad \\
\texttt{invest} & start a facility upgrade: academy, training, or
stadium \\
\texttt{set\_standing\_order} & add or clear a standing order (max 10;
one threshold plus a fixed action) \\
\texttt{submit\_draft} & draft phase only: submit the complete pick
list; resubmitting replaces it, and auto-fill completes a short list
from the cheapest tier \\
\midrule
\multicolumn{2}{@{}l}{\emph{Notebook (2), free:}} \\
\texttt{append\_note} & append to the private notebook (persists across
stops) \\
\texttt{rewrite\_notes} & replace the entire notebook \\
\midrule
\multicolumn{2}{@{}l}{\emph{Control (1):}} \\
\texttt{advance} & end this decision stop and simulate to the next
one \\
\bottomrule
\end{tabularx}
\end{table}

\begin{table}[t]
\centering
\small
\caption{The decision-stop calendar: 13 scheduled stops per season on
the 360-day year, plus event-driven stops the world raises unsolicited.
Several types can fire at one stop.}
\label{tab:stops}
\begin{tabularx}{\textwidth}{@{}>{\raggedright\arraybackslash}p{4.6cm}X@{}}
\toprule
stop (day of the 360-day year) & what the world wants decided \\
\midrule
\multicolumn{2}{@{}l}{\emph{Scheduled, 13 per season:}} \\
Preseason board (2) & the board sets the season target; plan the year \\
Summer window plan (5) & the transfer window opens; build the squad \\
Summer deadline (59) & last actions before the window closes \\
Lineup lock (63) & commit lineup and tactics before round 1 \\
Monthly report (90, 120, 150, 240) & finances and form check-in \\
Winter window open (181) & mid-course squad correction \\
Midseason review (190) & the board measures progress against target \\
Winter deadline (209) & last winter-window actions \\
Season settlement (300) & the season closes: honors, revenue, board
verdict \\
Youth intake (305) & the academy class arrives; promote, hold, or
release \\
\midrule
\multicolumn{2}{@{}l}{\emph{Event-driven, unsolicited:}} \\
Draft (once, at run start) & assemble the 25-player squad from the
shared pool \\
Offer batch & incoming bids for the agent's players, batched at a
minimum 15-day spacing \\
Contract expiry & a senior contract is running down; renew or lose the
player for free \\
Major injury & a first-team player goes down; cover or reshuffle \\
Board warning & confidence is deteriorating; the board expects a
response \\
Insolvency warning & cash runway is shrinking (30- and 60-day warnings) \\
Administration & insolvency executed: points penalty and forced sales \\
Revival & the seat restarts under the capped-revival mechanism
(\S\ref{sec:campaign}) \\
\bottomrule
\end{tabularx}
\end{table}

\begin{table}[t]
\centering
\small
\caption{The four memory layers (\S\ref{sec:memory}). \emph{What
happened} is recorded automatically; \emph{what it means and what to do
next} must be written by the agent itself.}
\label{tab:memory}
\begin{tabularx}{\textwidth}{@{}>{\raggedright\arraybackslash}p{2.4cm}>{\raggedright\arraybackslash}p{1.6cm}X>{\raggedright\arraybackslash}p{3.9cm}@{}}
\toprule
layer & who writes it & contents & delivery \\
\midrule
Packet auto-context & engine & dashboard digest, inbox (all events since last
stop), agent's last 10 engine-mutating actions, pending offers & pushed free
into every stop packet \\
Archive & engine & honors / season / transfer history of the whole run &
\texttt{get\_history} query tool, on demand \\
Notebook & \textbf{the agent} & intentions, reasoning, plans (``why I bought
him'', ``sell Y in winter'') & injected into every packet; written via
\texttt{append\_note} / \texttt{rewrite\_notes} \\
Audit log & runner & every successful action &
\textbf{never fed back}: replay/anti-cheat/resume only \\
\bottomrule
\end{tabularx}
\end{table}

The scale of a single 20-year episode, measured over the solo track (Table~\ref{tab:scale}). How many turns a stop takes, how much a model reads before acting, and how its notebook grows are all the model's own choices; the spread in each row is behavioral, not a harness setting (\S\ref{sec:capabilities}). Model turns, and state-changing actions are deliberately distinct counts: a turn may only read or deliberate without touching the world, and a single turn may commit several actions at once, so the two layers separate thinking from deciding.

\begin{table}[t]
\centering
\small
\caption{One 20-year run, by the numbers: medians and ranges
across the 15 solo models.}
\label{tab:scale}
\begin{tabular}{@{}lll@{}}
\toprule
quantity & median & range \\
\midrule
decision stops & 374 & 341--390 \\
\quad of which scheduled & ${\sim}260$ (13/season) &
  rest event-triggered (Table~\ref{tab:stops}) \\
matches simulated in the world & 4{,}800 &
  (30 rounds $\times$ 8 matches $\times$ 20 seasons) \\
model turns (API calls) & ${\sim}1{,}760$ & 1{,}184--6{,}969 \\
state-changing actions & --- & 849--2{,}198 \\
tokens per run (manifest accounting) & 61M & 24M--274M across single runs \\
notebook writes & 354 & 320--425 \\
notebook size at year 20 & --- & 0.4k--209k characters \\
simulation parameters (frozen) & 313 & --- \\
\bottomrule
\end{tabular}
\end{table}

\section{One Decision Stop, Concretely}\label{app:turn}

A slice of one transfer-window stop as the agent sees it (abridged from the public leaderboard page; the full rules document ships with the repository):

\begin{verbatim}
<- the game shows the current state (abridged)
{ "stop": "transfer_window", "date": "Y12-D181",
  "cash_m": 240.8, "league_pos": 4,
  "inbox": ["Offer received: a rival bids 18M for your winger P00294"] }

-> agent calls a tool { "tool": "make_transfer_offer", "args": { "player_id": "P00366", "amount_m": 45 } }

<- engine responds { "ok": true, "note": "sealed bid lodged; resolves at day end" }

-> agent keeps acting in the same stop...
{ "tool": "offer_contract",
  "args": { "player_id": "P00366", "wage_m_per_year": 15, "years": 5 } }
{ "tool": "advance", "args": {} }   // done; the world resumes
\end{verbatim}

Within a stop, the agent may take as many turns as it wants (query budgets permitting, \S\ref{sec:game}); \texttt{advance} terminates the stop, and the world simulates day-by-day to the next one. Between stops, standing intent (lineups, tactics, standing orders, in-flight negotiations under TTL) remains in force.

\end{document}